\documentclass[11pt,a4paper]{article}

\usepackage[utf8]{inputenc}
\usepackage[T1]{fontenc}
\usepackage{mathpazo}               % Palatino font (matches MDPI style)
\usepackage{amsmath,amssymb,amsfonts}
\usepackage{graphicx}
\usepackage{booktabs}
\usepackage{tabularx}
\usepackage{float}
\usepackage{hyperref}
\usepackage{url}
\usepackage{geometry}
\usepackage{authblk}
\usepackage{lineno}
\usepackage{tikz}
\usetikzlibrary{arrows.meta,positioning,shapes.geometric,fit,calc}
\usepackage{changepage}             % for adjustwidth if needed
\usepackage{xcolor}

\newcolumntype{C}{>{\centering\arraybackslash}X}

\hypersetup{
    colorlinks=true,
    linkcolor=blue!60!black,
    citecolor=blue!60!black,
    urlcolor=blue!60!black,
}

\title{\textbf{ElephantBroker: A Knowledge-Grounded Cognitive Runtime\\for Trustworthy AI Agents}}

\author[1]{Cristian Lupascu, PhD\thanks{Correspondence: cristian@elephant.broker}}
\author[1]{Alexandru Lupascu}
\affil[1]{Elephant Broker, 050141 Bucharest, Romania}

\date{}

\begin{document}

\maketitle

%=================================================================
% Abstract
%=================================================================
\begin{abstract}
Large Language Model (LLM)-based agents increasingly operate in high-stakes, multi-turn settings where factual grounding is critical, yet their memory systems typically rely on flat key--value stores or plain vector retrieval with no mechanism to track the provenance or trustworthiness of stored knowledge. We present ElephantBroker, an open-source cognitive runtime that unifies a Neo4j knowledge graph with a Qdrant vector store through the Cognee SDK to provide durable, verifiable agent memory. The system implements a complete cognitive loop (store, retrieve, score, compose, protect, learn) comprising a hybrid five-source retrieval pipeline, an eleven-dimension competitive scoring engine for budget-constrained context assembly, a four-state evidence verification model, a five-stage context lifecycle with goal-aware assembly and continuous compaction, a six-layer cheap-first guard pipeline for safety enforcement, an AI firewall providing enforceable tool-call interception and multi-tier safety scanning, a nine-stage consolidation engine that strengthens useful patterns while decaying noise, and a numeric authority model governing multi-organization identity with hierarchical access control. Architectural validation through a comprehensive test suite of over 2{,}200 tests spanning unit, integration, and end-to-end levels confirms subsystem correctness. The modular design supports three deployment tiers, five profile presets with inheritance, multi-gateway isolation, and a management dashboard for human oversight, enabling configurations from lightweight memory-only agents to full cognitive runtimes with enterprise-grade safety and auditability.
\end{abstract}


\vspace{1em}

%%%%%%%%%%%%%%%%%%%%%%%%%%%%%%%%%%%%%%%%%%
\section{Introduction}
\label{sec:introduction}

The rapid adoption of Large Language Model (LLM)-based agents in domains such as customer support, scientific research, and enterprise automation has created a new class of software systems that must maintain coherent memory across extended interactions~\cite{ref-park2023,ref-sumers2024}. Unlike traditional stateless question-answering pipelines, these agents accumulate knowledge over dozens or hundreds of conversational turns, make decisions based on previously stored facts, and operate within strict context-window budgets imposed by the underlying language models.

This shift from stateless to stateful AI systems raises a fundamental question: how should an agent decide which pieces of its accumulated knowledge deserve to occupy the scarce real estate of the context window? The answer has profound implications for trustworthiness. An agent that fills its context with unverified hearsay while leaving tool-verified facts behind will inevitably produce less reliable outputs, regardless of how capable the underlying language model may be.

\subsection{Motivation}
\label{subsec:motivation}

Contemporary agent frameworks typically store memories as flat text entries in a vector database and retrieve them through embedding similarity at query time~\cite{ref-lewis2020}. While this approach has proven effective for single-turn retrieval-augmented generation (RAG), it suffers from several limitations when applied to persistent agent memory.

First, vector similarity alone cannot capture the rich relational structure that emerges in multi-turn conversations. When an agent learns that ``Alice manages the Berlin office'' and later discovers that ``the Berlin office handles European compliance,'' a pure vector store cannot traverse the implicit relationship between Alice and European compliance without an explicit graph representation. Second, all retrieved memories are treated as equally trustworthy, regardless of whether they originated from a verified tool output, an unconfirmed user statement, or a superseded earlier assertion. Third, the retrieval process is typically disconnected from the agent's current goals, meaning that goal-relevant memories compete on equal footing with tangentially related but higher-similarity entries.

These shortcomings become critical in high-stakes domains. A medical assistant that retrieves a contradictory drug interaction from memory, or a legal research agent that cannot distinguish verified precedent from unverified conjecture, may cause material harm~\cite{ref-ji2023}. The problem is compounded in multi-agent settings where users, sub-agents, and supervisors contribute knowledge with differing authority levels~\cite{ref-hong2024}.

\subsection{Current Approaches and Their Limitations}
\label{subsec:current}

Several recent systems have attempted to address parts of this problem. MemGPT~\cite{ref-packer2024} introduces an operating-system-inspired virtual memory hierarchy that pages information in and out of the LLM context, but treats memory pages as opaque text blocks without relational structure or provenance tracking. Generative Agents~\cite{ref-park2023} maintains episodic memory streams that support reflection and planning, demonstrating the value of structured memory for believable agent behavior, yet stores memories as flat entries without evidence tracking or goal-aware prioritization.

On the retrieval side, standard RAG pipelines~\cite{ref-lewis2020,ref-gao2024} condition generation on passages retrieved from external corpora but operate statelessly, rebuilding context from scratch at each query rather than maintaining persistent memory. Graph-enhanced approaches such as GraphRAG~\cite{ref-edge2024} build knowledge graphs from document corpora for query-time summarization, yet these graphs are typically constructed offline from static documents rather than incrementally from live conversations.

The integration of knowledge graphs with LLMs has been explored for factual grounding~\cite{ref-pan2024} and hallucination detection~\cite{ref-manakul2023}, but existing systems treat the knowledge graph as a read-only reference rather than a dynamic memory that evolves with each agent interaction. None of the current approaches combine persistent graph-structured memory, evidence-tracked verification, goal-aware retrieval, budget-constrained context assembly, safety guard pipelines, and consolidation-based learning into a unified runtime.

\subsection{Research Contribution}
\label{subsec:contribution}

In this paper, we present ElephantBroker, an open-source cognitive runtime that addresses the aforementioned limitations through eight interconnected mechanisms. First, a \textbf{hybrid retrieval pipeline} that concurrently queries five complementary sources (structural graph queries, lexical matching, semantic vector search, ego-net graph expansion, and artifact retrieval) and merges results through configurable per-source weights with actor-level isolation policies. Second, an \textbf{eleven-dimension scoring engine} that evaluates each retrieved candidate across turn relevance, goal relevance (both session and global), recency, use history, confidence with verification multipliers, evidence strength, novelty, redundancy, contradiction, and token cost, then assembles a working set through budget-constrained greedy selection with a $(1-1/e)$ submodularity guarantee~\cite{ref-nemhauser1978}. Third, a \textbf{four-state evidence verification model} where stored facts accumulate typed evidence over time, and their verification status directly influences scoring priority, creating a structural bias toward trustworthy information. Fourth, a \textbf{five-stage context lifecycle} (bootstrap, ingest, assemble, compact, afterTurn) that manages profile resolution, four-block assembly across two prompt surfaces, two-stage compaction, and six-signal successful-use detection, with subagent spawn and end hooks for delegated workflows. Fifth, a \textbf{six-layer cheap-first guard pipeline} that enforces safety through autonomy domain classification, static rule matching, semantic similarity, structural validation, forced constraint reinjection, and optional LLM escalation, with procedure-bound red-line constraints that dynamically compose the active rule set based on which workflows are running and near-miss tracking that auto-tightens strictness under boundary-probing behavior. Sixth, a \textbf{nine-stage consolidation engine} that runs during idle periods to cluster near-duplicates, canonicalize, strengthen useful facts, decay unused ones, prune ineffective auto-recall, promote recurring episodic knowledge to semantic, refine procedures from cross-session patterns, identify verification gaps, and recompute salience priors. Seventh, an \textbf{AI firewall} that transforms advisory guard verdicts into physically enforceable tool-call interception across three defense layers (agent hook, proxy guardrails, SDK hooks), with cost-tiered safety scanning that detects prompt injection, jailbreak attempts, PII leakage, and toxic content through a six-tier pipeline from regex patterns through fine-tuned classification models, and prevents knowledge base contamination by blocking adversarial content at the ingest boundary. Eighth, an \textbf{authority-based multi-organization identity model} implementing a numeric permission hierarchy with four authority tiers (regular actor, team lead, organization admin, system admin), eleven configurable authority rules with matching-exempt semantics analogous to Unix privilege escalation, organization and team entities as first-class graph nodes, platform-qualified actor handles, scope-aware goal visibility, and a management dashboard for human oversight.

The runtime is built as a Python-native service backed by Neo4j, Qdrant, and the Cognee SDK~\cite{ref-cognee}, integrated with OpenClaw, an open-source TypeScript runtime for building LLM-based agents, through thin TypeScript plugin adapters that expose twenty-four tools covering memory operations, session goal management, procedure lifecycle tracking, artifact handling, guard inspection, and administrative operations.
In the next section, we review the background technologies and relevant related work. In Section~\ref{sec:proposed}, we describe the proposed cognitive runtime architecture in detail. We present the validation results in Section~\ref{sec:results} and conclude the paper in Section~\ref{sec:conclusions}.

%%%%%%%%%%%%%%%%%%%%%%%%%%%%%%%%%%%%%%%%%%
\section{Background and Related Work}
\label{sec:background}

\subsection{Knowledge Graphs for Language Model Grounding}
\label{subsec:kg}

Knowledge graphs represent factual knowledge as directed labeled graphs where nodes denote entities and edges denote relationships. In the context of LLM-based systems, knowledge graphs have been used to reduce hallucination by providing structured factual grounding~\cite{ref-pan2024}. The key advantage over flat text retrieval is the ability to traverse multi-hop relationships: given a query about European compliance policies, a graph traversal can follow edges from a compliance node through office and manager relationships to surface relevant personnel, a connection that vector similarity alone would miss.

Neo4j, the graph database used in our implementation, stores data as property graphs where both nodes and edges carry arbitrary key--value attributes. Its Cypher query language enables expressive pattern matching that we exploit for structural retrieval, for example querying all facts created by a specific actor within a particular session scope.

The Cognee SDK~\cite{ref-cognee} (used in our implementation) provides a higher-level abstraction over graph and vector storage through its DataPoint model. The SDK's storage API atomically creates graph nodes via merge operations and vector embeddings via configurable index fields in a single call, ensuring consistency between the two storage backends. This dual-write architecture means that every stored fact is simultaneously accessible through graph traversal and vector similarity, a property we rely on heavily in our hybrid retrieval pipeline. Additionally, the SDK's enrichment pipeline performs entity extraction, relationship extraction, and triplet embedding, building the rich graph indexes that power multi-hop retrieval. Luo et al.~\cite{ref-luo2024} demonstrate that graph-based reasoning produces more faithful and interpretable LLM outputs compared to text-only retrieval, reinforcing the case for graph-structured memory in agent systems.

\subsection{Retrieval-Augmented Generation}
\label{subsec:rag}

Retrieval-Augmented Generation (RAG)~\cite{ref-lewis2020} established the paradigm of conditioning LLM outputs on retrieved passages from external corpora. The approach has since evolved along several axes: dense--sparse hybrid retrieval~\cite{ref-ma2024} combines embedding similarity with lexical matching for more robust recall; multi-hop retrieval~\cite{ref-trivedi2023} chains multiple retrieval steps to answer complex questions; and self-reflective RAG~\cite{ref-asai2024} introduces verification loops where the model critiques its own retrieved context.

GraphRAG~\cite{ref-edge2024} represents a significant advance by building knowledge graphs from document corpora and using community-based summarization to support global queries that span multiple topics. However, GraphRAG constructs its graphs offline through batch processing of static documents, whereas our system builds its knowledge graph incrementally from live conversational turns, with each interaction potentially adding facts, updating confidence scores, or providing new evidence.

\subsection{Agent Memory and Cognitive Architectures}
\label{subsec:memory}

The design of agent memory systems draws on a long tradition in cognitive architectures. Soar~\cite{ref-laird2012} and ACT-R~\cite{ref-anderson2004} introduced the distinction between working memory (actively maintained context), episodic memory (records of experiences), semantic memory (general knowledge), and procedural memory (skills and procedures). Modern LLM-based agents have begun to adopt similar taxonomies~\cite{ref-sumers2024}, though typically with simpler implementations.

Generative Agents~\cite{ref-park2023} demonstrated that maintaining an episodic memory stream with importance scoring and periodic reflection produces more believable agent behavior than stateless generation. MemGPT~\cite{ref-packer2024} introduced the analogy between an operating system's virtual memory and the LLM's context window, with explicit page-in and page-out operations. More recently, cognitive architecture surveys~\cite{ref-sumers2024} have called for unified frameworks that combine retrieval, reasoning, and memory management, precisely the gap that ElephantBroker aims to fill. Reflexion~\cite{ref-shinn2023} demonstrates that verbal self-evaluation loops improve agent task completion, a principle that our consolidation engine extends from single-task reflection to cross-session memory refinement.

\subsection{Contemporary Agent Memory Systems}
\label{subsec:contemporary}

The practical landscape of agent memory has grown rapidly, with systems targeting different points in the design space. Zep~\cite{ref-zep2025} introduces Graphiti, a temporally-aware knowledge graph engine that organizes memory into three hierarchical tiers: episodic nodes (raw messages), semantic entities and facts (extracted knowledge with bi-temporal edge validity), and community summaries (cluster-level abstractions derived via label propagation). Zep's retrieval pipeline composes three search methods (cosine similarity, BM25 full-text, and breadth-first graph traversal) with reranking (including reciprocal rank fusion, maximal marginal relevance, and optional cross-encoder scoring) and a context constructor that formats facts with their temporal validity ranges. Both systems build knowledge graphs incrementally from conversations and support bi-temporal reasoning: Zep tracks four timestamps per edge ($t'_{\text{created}}$, $t'_{\text{expired}}$ on the transactional timeline and $t_{\text{valid}}$, $t_{\text{invalid}}$ on the real-world timeline), while ElephantBroker stores event timestamps, ingestion timestamps, and edge validity intervals to enable point-in-time knowledge graph queries. The key architectural difference is that Zep optimizes for retrieval accuracy and latency, achieving state-of-the-art results on the Deep Memory Retrieval~\cite{ref-packer2024} and LongMemEval~\cite{ref-longmemeval} benchmarks, while ElephantBroker extends the retrieval foundation with evidence verification, multi-dimensional scoring with budget competition, goal-aware context assembly, a guard pipeline for safety enforcement, and consolidation-based learning.

Mem0~\cite{ref-mem0-2025} provides a scalable memory-centric architecture that extracts, consolidates, and retrieves salient information from ongoing conversations, with an optional graph memory variant for capturing relational structures. Mem0 achieves strong retrieval performance and low latency but treats all memories as equally trustworthy and does not provide goal-aware scoring or budget-constrained working set assembly.

Several other systems address narrower aspects of the agent memory problem. Lossless-Claw~\cite{ref-losslessclaw} focuses exclusively on within-session lossless context compression through a directed acyclic graph (DAG) representation, preserving all information from the current conversation without cross-session persistence. QMD~\cite{ref-qmd} provides a local hybrid search sidecar that runs entirely on-device with local embeddings, prioritizing data sovereignty and low latency over relational knowledge representation. Supermemory~\cite{ref-supermemory} offers a cloud-hosted memory API with atomic trace storage and profile-based organization, trading data locality for operational simplicity.

Table~\ref{tab:comparison} summarizes how ElephantBroker relates to these systems across key architectural dimensions.

\begin{table}[H]
\caption{Comparison of agent memory systems across key architectural dimensions.\label{tab:comparison}}
\centering
\setlength{\tabcolsep}{3pt}
\begin{tabularx}{\textwidth}{lCCCCCCC}
\toprule
\textbf{Capability} & \textbf{Ours} & \textbf{Zep} & \textbf{Mem0} & \textbf{Mem\-GPT} & \textbf{L.C.} & \textbf{QMD} & \textbf{S.M.} \\
\midrule
Knowledge graph & \checkmark & \checkmark & Opt. & --- & --- & --- & --- \\
Vector store & \checkmark & \checkmark & \checkmark & --- & --- & \checkmark & \checkmark \\
Evidence tracking & \checkmark & --- & --- & --- & --- & --- & --- \\
Goal-aware scoring & \checkmark & --- & --- & --- & --- & --- & --- \\
Working set budget & \checkmark & --- & --- & \checkmark & --- & --- & --- \\
Multi-agent actors & \checkmark & --- & Part. & --- & --- & --- & Part. \\
Cross-session mem. & \checkmark & \checkmark & \checkmark & \checkmark & --- & \checkmark & \checkmark \\
Temporal reasoning & \checkmark & \checkmark & --- & --- & --- & --- & --- \\
Session goals & \checkmark & --- & --- & --- & --- & --- & --- \\
Procedure tracking & \checkmark & --- & --- & --- & --- & --- & --- \\
Guard pipeline & \checkmark & --- & --- & --- & --- & --- & --- \\
Consolidation & \checkmark & --- & Part. & --- & --- & --- & --- \\
Authority model & \checkmark & --- & --- & --- & --- & --- & --- \\
Enforceable guards & \checkmark & --- & --- & --- & --- & --- & --- \\
AI safety scanning & \checkmark & --- & --- & --- & --- & --- & --- \\
Mgmt.\ dashboard & \checkmark & --- & --- & --- & --- & --- & --- \\
\bottomrule
\end{tabularx}
\end{table}

\subsection{Trustworthy AI and Verification}
\label{subsec:trustworthy}

Trustworthiness in AI systems encompasses factuality, faithfulness, calibration, and alignment~\cite{ref-ji2023,ref-huang2023}. Approaches to hallucination mitigation include post-hoc verification against external sources~\cite{ref-min2023}, chain-of-verification prompting~\cite{ref-dhuliawala2023}, and confidence calibration~\cite{ref-kadavath2022}. However, these methods operate at inference time without maintaining a persistent verification state across sessions.

From a regulatory perspective, the EU AI Act~\cite{ref-euaiact} requires high-risk AI systems to implement risk management, data governance, transparency, and human oversight. ElephantBroker's evidence verification model provides a mechanism for several of these requirements: the supervisor verification state enables human-in-the-loop oversight, the evidence tracking supports auditability, the trace ledger provides transparency through a comprehensive audit trail, the GDPR-compliant forget operation with data-minimized audit events supports the right to erasure, and over 100 Prometheus metrics covering retrieval performance, scoring decisions, guard outcomes, and goal lifecycle events provide the operational transparency needed for ongoing risk monitoring. A web-based management dashboard further supports the Article~14 requirement for ``effective oversight by natural persons'' through a memory browser, guard event review, and cross-session approval queue~\cite{ref-nist-ai}.

\subsection{LLM Safety and Guardrails}
\label{subsec:safety}

Indirect prompt injection, where adversarial instructions are embedded in retrieved context or tool outputs, poses a fundamental threat to agent systems that consume external content~\cite{ref-greshake2023}. The OWASP Foundation identifies prompt injection as the highest-severity risk in its Top~10 for LLM Applications~\cite{ref-owasp-llm}, and large-scale red-teaming has exposed systematic weaknesses across production language models~\cite{ref-perez2023}. These attacks are particularly dangerous in persistent memory systems because adversarial content stored as a ``fact'' can persist across sessions and influence future agent behavior through the retrieval pipeline.

Several frameworks address LLM safety at different layers. LLM~Guard~\cite{ref-llmguard} provides input/output scanners for PII detection, toxicity filtering, and prompt injection detection using fine-tuned classification models. Guardrails~AI~\cite{ref-guardrailsai} offers a declarative specification language for defining structural and content constraints on LLM outputs with automatic retry and correction. NeMo Guardrails~\cite{ref-nemo} implements programmable conversation rails that intercept and redirect LLM interactions at the dialog level. Rebuff~\cite{ref-rebuff} provides a multi-layer prompt injection defense combining heuristic analysis, LLM-based detection, and canary token injection. LiteLLM~\cite{ref-litellm} provides a unified proxy layer with pluggable guardrails hooks that can intercept both prompts and completions. A comprehensive survey of LLM safeguarding approaches is provided by Dong et al.~\cite{ref-dong2024}.

While these frameworks address individual safety concerns effectively, none integrates safety scanning with a persistent memory system's guard pipeline, evidence model, and autonomy classification. A prompt injection detected in a tool output should not only block the immediate interaction but also prevent the malicious content from being stored as a trusted fact in the knowledge graph. ElephantBroker's AI Firewall (Section~\ref{subsec:firewall}) addresses this gap by layering safety scanning on top of the existing guard pipeline and connecting detection events to the evidence and trace infrastructure.

%%%%%%%%%%%%%%%%%%%%%%%%%%%%%%%%%%%%%%%%%%
\section{Proposed Cognitive Runtime}
\label{sec:proposed}

ElephantBroker implements a complete cognitive loop for LLM-based agents: \textit{store} knowledge with typed provenance, \textit{retrieve} it through multi-source hybrid search, \textit{score} candidates in a budget competition weighted by eleven dimensions, \textit{compose} context through four-block assembly across two prompt surfaces, \textit{protect} actions through a six-layer guard pipeline, \textit{enforce} safety through physical tool-call interception, and \textit{learn} from usage patterns through nine-stage consolidation. The following subsections describe each component in detail.

\subsection{Architecture Overview}
\label{subsec:architecture}

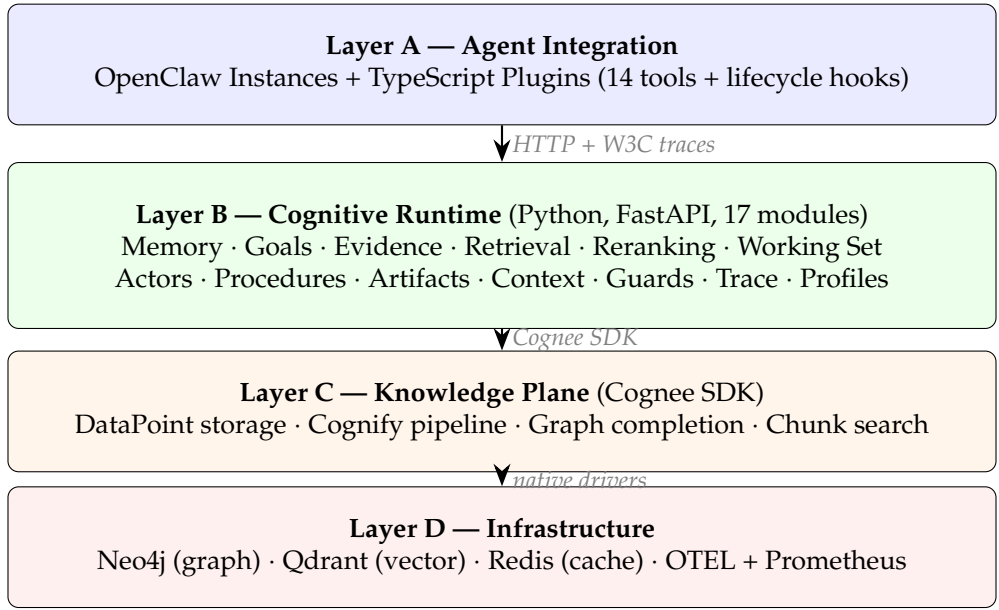
\begin{figure}[H]
\centering
\begin{tikzpicture}[
    layer/.style={draw, rounded corners, minimum width=13cm, minimum height=1.6cm, align=center, font=\small},
    arrow/.style={-{Stealth[length=3mm]}, thick},
    lbl/.style={font=\footnotesize\itshape, text=gray}
]
\node[layer, fill=blue!8] (A) at (0,6) {\textbf{Layer A --- Agent Integration}\\OpenClaw Instances + TypeScript Plugins (14 tools + lifecycle hooks)};
\node[layer, fill=green!8, minimum height=2.2cm] (B) at (0,3.6) {\textbf{Layer B --- Cognitive Runtime} (Python, FastAPI, 17 modules)\\Memory $\cdot$ Goals $\cdot$ Evidence $\cdot$ Retrieval $\cdot$ Reranking $\cdot$ Working Set\\Actors $\cdot$ Procedures $\cdot$ Artifacts $\cdot$ Context $\cdot$ Guards $\cdot$ Trace $\cdot$ Profiles};
\node[layer, fill=orange!8] (C) at (0,1.4) {\textbf{Layer C --- Knowledge Plane} (Cognee SDK)\\DataPoint storage $\cdot$ Cognify pipeline $\cdot$ Graph completion $\cdot$ Chunk search};
\node[layer, fill=red!6] (D) at (0,-0.4) {\textbf{Layer D --- Infrastructure}\\Neo4j (graph) $\cdot$ Qdrant (vector) $\cdot$ Redis (cache) $\cdot$ OTEL + Prometheus};

\draw[arrow] (A.south) -- (B.north) node[midway, right, lbl] {HTTP + W3C traces};
\draw[arrow] (B.south) -- (C.north) node[midway, right, lbl] {Cognee SDK};
\draw[arrow] (C.south) -- (D.north) node[midway, right, lbl] {native drivers};
\end{tikzpicture}
\caption{Four-layer architecture of ElephantBroker. Thin TypeScript adapters in Layer~A forward requests via HTTP to the Python runtime in Layer~B, which delegates storage and search to the Cognee knowledge plane in Layer~C, backed by the infrastructure services in Layer~D.\label{fig:architecture}}
\end{figure}

ElephantBroker is organized in four layers, as shown in Figure~\ref{fig:architecture}. Layer~A consists of thin TypeScript plugins that integrate with the OpenClaw agent platform, exposing twenty-four tools organized in six groups: five for memory operations (search, get, store, forget, update), five for session goal management (list, create, update status, add blocker, record progress), four for procedure lifecycle tracking (create, activate, complete step, session status), two for artifact handling (search, create), two for guard inspection (list rules, check status), and six for administrative operations (organization, team, actor, member, profile override, and actor merge management), along with lifecycle hooks for automatic memory capture at session boundaries. These plugins communicate via HTTP with W3C trace propagation to Layer~B, the Python-based cognitive runtime. Layer~B implements 17 modules (actor registry, goal manager, memory store facade, working set manager, context assembler, compaction engine, procedure engine, evidence and verification engine, guard engine, tool artifact store, retrieval orchestrator, rerank orchestrator, statistics engine, consolidation engine, profile registry, trace ledger, and scoring tuner) wired together by a dependency injection container that instantiates only the modules enabled by the selected business tier, ensuring that lightweight deployments do not pay the cost of unused subsystems. Layer~C is the Cognee knowledge plane, which provides unified graph--vector storage and search through its DataPoint abstraction and automated indexing pipelines. Layer~D comprises the infrastructure services: Neo4j for graph storage, Qdrant for vector search with dense embeddings, Redis for caching and session goals, and OpenTelemetry with Prometheus for observability across over 100 custom metrics covering store operations, retrieval performance, scoring pipeline latency, embedding cache hit rates, guard outcomes, consolidation progress, and goal and procedure lifecycle events. Both the LLM and the embedding model are configurable via a proxy layer, allowing operators to swap providers without code changes.

The plugin architecture is designed for extensibility. The TypeScript plugins are thin HTTP clients with no business logic; all intelligence resides in the Python runtime, making it straightforward to adapt the system to agent frameworks beyond OpenClaw by implementing the same HTTP contract. The twenty-four tools and lifecycle hooks are independently registrable, enabling partial adoption: an agent framework can integrate only the memory tools without goal management, or adopt goal tracking without the full context lifecycle. Custom guard rules, profile overrides, and procedure definitions can be injected at runtime through the API without redeployment, and the profile inheritance chain (base $\to$ preset $\to$ organization override) allows enterprise teams to customize behavior without forking the core configuration.

The system uses a three-level identity model. A \textbf{gateway identifier} uniquely names each OpenClaw deployment instance and serves as the top-level isolation boundary for cache key namespaces, graph queries, vector dataset names, metric labels, trace enrichment, and log prefixes. Below the gateway, each \textbf{agent} is identified by a deterministic UUID~v5 derived from the composite gateway--agent key, ensuring a stable identity across restarts. The agent registers as an actor reference on first session start, enabling proper per-message fact attribution: assistant and tool messages are attributed to the agent actor, while user messages are attributed to the speaker actor. Below the agent, \textbf{sessions} are identified by a stable routing key and an ephemeral session UUID that changes on reset. Multiple sessions share one agent identity, including subagent sessions, which do not create new agent identities. The TypeScript plugins transmit four identity headers on every HTTP request, extracted by gateway identity middleware. Missing gateway identifiers are rejected. Optional organization and team identifiers provide business-level isolation for profile overrides, consolidation scoping, and goal visibility, independent of infrastructure-level gateway isolation.

\subsection{Data Model and Actor Hierarchy}
\label{subsec:datamodel}

The system's knowledge base is modeled as a property graph $\mathcal{G} = (\mathcal{N}, \mathcal{E}, \phi)$, where $\mathcal{N}$ denotes nodes (facts, actors, goals, procedures, claims, evidence, artifacts), $\mathcal{E}$ denotes typed directed edges (authorship, supersession, contradiction, goal-serving, ownership, parent--child), and $\phi : \mathcal{E} \to \mathcal{N} \times \mathcal{N}$ is the incidence function. Each node and edge carries a property map of key--value attributes. The graph is paired with a vector index $\mathcal{V}$ that stores dense embeddings for designated text fields, enabling both structural traversal over $\mathcal{G}$ and semantic similarity search over $\mathcal{V}$ from a single storage operation.

The core entity is the \textit{fact assertion}, which represents a unit of agent memory and includes fields for content text, a category drawn from twelve built-in types (identity, preference, event, decision, system, relationship, trait, project, general, constraint, procedure reference, and verification) with support for custom category strings, a scope, a memory class, a confidence score, source and target actors, goal associations, and goal relevance tags. Pydantic~v2 schema classes define the full data model and serve as the single source of truth across all layers.

\subsubsection{Scope System}

The scope governs a fact's visibility, consolidation behavior, and promotion trajectory. Table~\ref{tab:scopes} lists all eight scope values with their respective retrieval visibility, decay cadence during consolidation, and the target scope upon promotion.

\begin{table}[H]
\caption{Eight-value scope system governing fact lifecycle.\label{tab:scopes}}
\centering
\begin{tabularx}{\textwidth}{lCCC}
\toprule
\textbf{Scope} & \textbf{Visibility} & \textbf{Decay Cadence} & \textbf{Promotion Target} \\
\midrule
GLOBAL & All gateways in org & Monthly & Terminal \\
ORGANIZATION & Matching org\_id & Weekly & GLOBAL \\
TEAM & Matching team\_id & Weekly & ORGANIZATION \\
ACTOR & Owner actor only & Daily & TEAM \\
SESSION & Current session & Hourly & ACTOR \\
TASK & Current task & Hourly & SESSION \\
SUBAGENT & Child session & Hourly & SESSION \\
ARTIFACT & Tool output ref & Daily & ACTOR \\
\bottomrule
\end{tabularx}
\end{table}

\subsubsection{Memory Classes}

Facts are classified into five memory classes following the cognitive science taxonomy. Each class exhibits distinct compaction behavior and consolidation treatment, as shown in Table~\ref{tab:memclass}.

\begin{table}[H]
\caption{Five memory classes with lifecycle properties.\label{tab:memclass}}
\centering
\begin{tabularx}{\textwidth}{lCCCC}
\toprule
\textbf{Class} & \textbf{Typical Content} & \textbf{Decay} & \textbf{Promotable} & \textbf{Compaction} \\
\midrule
EPISODIC & Events, decisions & Yes & SEMANTIC (3+ sessions) & Summarizable \\
SEMANTIC & Durable knowledge & Slow & Terminal & Mergeable \\
PROCEDURAL & Workflows, SOPs & None & Terminal & Preserve \\
POLICY & Constraints, rules & None & Terminal & Always inject \\
WORKING\_MEMORY & Transient context & Session-only & --- & Drop at session end \\
\bottomrule
\end{tabularx}
\end{table}

Classification uses a fast rule table at ingest time: constraint and procedure reference categories map to POLICY; preferences, identity, traits, relationships, projects, and system facts map to SEMANTIC; events, decisions, and verification results map to EPISODIC. Ambiguous cases are resolved by an LLM fallback. A second-pass reclassification during consolidation promotes recurring EPISODIC facts to SEMANTIC based on cross-session recurrence (three or more sessions).

\subsubsection{Actor Model}

Unlike conventional single-user memory systems, ElephantBroker implements a hierarchical actor model that supports multi-agent organizations. The system defines twelve actor types spanning human roles (coordinator, operator, external), agent roles (manager, worker, reviewer, supervisor, peer, service, external), and organizational units (teams and organizations). These actors form authority chains through twelve relationship types (delegates\_to, supervises, reports\_to, collaborates\_with, trusts, blocks, owns\_goal, owns\_artifact, requested\_by, approved\_by, verified\_by, and prohibited\_by) with configurable delegation relationships. Trust levels (0--1) on actor references affect evidence weighting in the scoring pipeline: a supervisor-verified claim from a high-trust actor produces a stronger confidence signal than one from a low-trust source. Actors support soft-deactivation via an active status field, preserving all audit trail entries and provenance edges while removing the actor from active queries and session access. The actor model uses typed graph edges (authorship, actor-about, goal-serving, goal-ownership) to maintain provenance chains that can be traversed during retrieval.

\subsubsection{DataPoint Subclasses}

Seven entity subclasses bridge the abstract data model to physical storage. Each subclass declares which fields are vector-indexed, as shown in Table~\ref{tab:datapoints}.

\begin{table}[H]
\caption{Seven entity subclasses mapping schema models to graph--vector storage.\label{tab:datapoints}}
\centering
\begin{tabularx}{\textwidth}{lClC}
\toprule
\textbf{Entity Type} & \textbf{Schema Source} & \textbf{Indexed Fields} & \textbf{Vector Collection} \\
\midrule
Fact & Fact assertion & text & Fact\_text \\
Actor & Actor reference & display name & Actor\_display\_name \\
Goal & Goal state & title, description & Goal\_title \\
Procedure & Procedure def. & name, description & Procedure\_name \\
Claim & Claim record & claim text & Claim\_claim\_text \\
Evidence & Evidence ref. & reference value & Evidence\_ref\_value \\
Artifact & Tool artifact & summary & Artifact\_summary \\
\bottomrule
\end{tabularx}
\end{table}

Storage follows a dual-write pattern. Step~1 atomically creates a graph node via merge-by-ID and indexes the declared fields in the vector store, providing immediate searchability. Step~2 runs the Cognee enrichment pipeline (entity extraction, relationship extraction, and triplet embedding) to build rich graph indexes that power multi-hop retrieval. Updates (confidence changes, scope promotions, status transitions) execute only Step~1, since the text is already queued from the original creation. Re-storing with the same ID triggers upsert semantics, enabling in-place updates without duplicates or orphaned graph edges.

\subsubsection{Temporal Reasoning}

Every fact and graph edge carries explicit temporal metadata that supports bi-temporal reasoning over the knowledge base. Each fact assertion records three timestamps: an \textit{event timestamp} (when the described event occurred in the real world), an \textit{ingestion timestamp} (when the system first stored the fact), and an \textit{update timestamp} (when any property was last modified). All typed graph edges (authorship, supersession, contradiction, goal-serving, goal-ownership, parent--child) carry \textit{valid-from} and \textit{valid-until} properties, enabling the system to reason about the temporal validity of relationships; for example, an actor's team membership that changed mid-session, or a goal hierarchy that was restructured. Supersession edges naturally encode temporal replacement: the old fact's valid-until is set to the new fact's ingestion timestamp, producing a non-overlapping validity chain. Similarly, contradiction edges record the moment of conflict detection, allowing the consolidation engine (Section~\ref{subsec:consolidation}) to distinguish long-standing contradictions from recently discovered ones. Structural retrieval queries can filter by temporal range (selecting edges whose valid-from precedes the query time and whose valid-until is either absent or follows the query time), enabling point-in-time knowledge graph snapshots. The recency scoring dimension (Equation~\ref{eq:recency}) leverages the last-used timestamp rather than the ingestion timestamp, ensuring that continuously useful facts are not penalized for age, while the consolidation engine's decay stage (Section~\ref{subsec:consolidation}, Stage~4) uses the ingestion timestamp for never-recalled facts to identify stale knowledge. All timestamps are stored as epoch milliseconds for efficient range queries and cross-timezone consistency.

\subsubsection{Isolation Model}

Memory isolation is configured along two axes. The \textit{isolation level} (NONE, LOOSE, STRICT) controls retrieval filtering: STRICT disables the keyword source (which can leak across scope boundaries) and uses direct vector queries with payload filters; LOOSE enables all five retrieval sources but applies post-filtering by session key or actor identity; NONE applies no filtering. The \textit{isolation scope} (GLOBAL, SESSION\_KEY, ACTOR, SUBAGENT\_INHERIT) determines the boundary for post-filtering. Per-profile defaults ensure appropriate isolation: personal assistant uses STRICT/SESSION\_KEY for privacy, research uses NONE/GLOBAL for maximum recall, and coding, managerial, and worker profiles use LOOSE/SESSION\_KEY.

\subsubsection{Deployment Tiers}

The runtime supports three deployment tiers that control which modules are instantiated: \textit{memory-only} (actors, goals, memory, retrieval, evidence, procedures), \textit{context-only} (actors, goals, working set, assembly, compaction, guards), and \textit{full} (all 17 modules). The dependency injection container instantiates only tier-enabled modules, with a tier manager that also supports granular feature flags (e.g., subagent context packets, consolidation, guard preflight checks) beyond module-level enablement. This tiered architecture ensures that a lightweight deployment paying only for durable memory does not incur the cost of context assembly or guard evaluation, while a full deployment gains the complete cognitive loop.

\subsection{Hybrid Five-Source Retrieval}
\label{subsec:retrieval}

A central design decision in ElephantBroker is that no single retrieval method is sufficient for the diversity of queries an agent encounters. A question about a specific user preference is best answered by a structural graph query that filters by actor and category. A question about a general concept benefits from semantic vector similarity. A question about relationships between entities requires graph traversal. We formalize the retrieval-to-context pipeline as a function composition $f(\alpha) = \sigma(\rho(\varphi(\alpha)))$, where $\alpha$ is the query string, $\varphi$ is the multi-source search function, $\rho$ is the four-stage reranker, and $\sigma$ is the eleven-dimension scoring and budget selection function. The output $f(\alpha)$ is a budget-constrained working set. This decomposition separates recall ($\varphi$), precision ($\rho$), and value optimization ($\sigma$) into independently tunable stages. Zep~\cite{ref-zep2025} employs a similar three-component decomposition (search, rerank, construct) but without the scoring stage, meaning all retrieved facts compete on relevance alone without considering evidence strength, goal alignment, or token cost.

To address query diversity, the search function $\varphi$ concurrently dispatches queries to five sources.

The \textit{structural source} executes graph pattern-matching queries against Neo4j, filtering by session key, actor, memory class, and scope. This source excels at precise property-based lookups; critically, the Cognee search API cannot filter by arbitrary node properties such as scope or actor identity, making structural graph queries essential. Results receive a default weight of $w_{\text{struct}} = 0.4$. The \textit{keyword source} performs lexical matching through Cognee's lexical chunk search, providing BM25-like recall for exact term matches with $w_{\text{kw}} = 0.3$. The \textit{semantic source} uses Cognee's semantic chunk search or direct vector queries for embedding-based similarity with $w_{\text{sem}} = 0.5$, falling back to direct vector search with payload filters under STRICT isolation. The \textit{graph expansion source} leverages Cognee's graph completion mode to traverse the ego-net around matching entities up to a configurable depth, discovering related facts through relationship traversal with $w_{\text{graph}} = 0.2$. The \textit{artifact source} searches tool output summaries indexed in dedicated vector collections with $w_{\text{art}} = 0.5$.

All five sources are dispatched concurrently. Upon completion, the orchestrator multiplies each candidate's raw score by its source weight and merges results by fact ID, retaining the maximum weighted score when a candidate appears in multiple sources. The merged list is deduplicated, sorted descending, and capped at a configurable maximum (configurable). All source weights, fetch counts, enable flags, graph mode (LOCAL, HYBRID, or GLOBAL), and graph depth are configurable per profile via the retrieval policy.

The pipeline enforces a five-level graceful degradation hierarchy. At the coarsest level, an entire source may be skipped (e.g., STRICT isolation disables keyword search). Finer-grained fallbacks include method-level degradation (Cognee search falls back to direct vector queries), pipeline-level degradation (reranker failure returns the input list unchanged), operation-level degradation (cache failure proceeds without caching), and edge-level degradation (graph edge creation failure logs a warning without blocking the store operation). This hierarchy ensures that no single backend failure blocks the pipeline, with each degradation event recorded for operational visibility.

\subsection{Four-Stage Reranking}
\label{subsec:reranking}

Before entering the scoring engine, retrieved candidates pass through a four-stage reranking pipeline designed to progressively refine the candidate set using increasingly expensive computations.

The first stage is a cheap prune that computes a blended score: 50\% token overlap between query and candidate text plus 50\% original retrieval score. This lightweight filter caps the candidate set at a configurable maximum, discarding clearly irrelevant matches without incurring embedding computation.

The second stage performs semantic reranking by computing cosine similarity between query and candidate embeddings and blending with the retrieval score using a configurable blend weight ($\alpha$).

The third stage invokes a configurable external cross-encoder model via an HTTP reranking endpoint. Raw logit margins are normalized through a sigmoid function. The stage processes candidates in configurable batches with a configurable timeout. This stage degrades gracefully: on timeout or endpoint failure, it returns the semantic reranking order and emits a degraded-operation trace event, ensuring that the overall pipeline never blocks on an unavailable reranker.

The fourth stage merges near-duplicate candidates using a union-find algorithm. Candidates whose embedding cosine similarity exceeds 0.95 are clustered, and within each cluster, only the highest-scored candidate is retained, with relations from all cluster members unioned onto the surviving candidate. This prevents the working set from being dominated by paraphrases of the same fact.

\subsection{Eleven-Dimension Competitive Scoring}
\label{subsec:scoring}

The scoring engine is the mechanism that transforms a ranked list of retrieved candidates into a budget-constrained working set. The key insight is that relevance alone is insufficient for trustworthy context assembly: the system must also consider recency, confidence, evidence support, goal alignment, novelty, and cost. We organize these concerns into eleven scoring dimensions computed in two passes.

\subsubsection{Scoring Context Pre-Computation}

Before scoring begins, the engine executes six parallel queries: (1)~evidence counts via a graph chain traversal (Evidence$\to$Claim$\to$Fact), (2)~verification statuses for all relevant claims, (3)~conflict pairs from supersession and contradiction graph edges, (4)~session goals from the cache, (5)~the current turn embedding, and (6)~persistent goals from the knowledge graph with active scopes. After the parallel phase, goal embeddings are computed sequentially via a single batch embedding call. This pre-computation avoids repeated per-candidate queries during the scoring loop.

\subsubsection{Pass 1: Independent Dimensions}

Nine dimensions are computed independently for each candidate. \textit{Turn relevance} is the cosine similarity between the candidate's embedding and the current conversational turn's embedding. \textit{Session goal relevance} first checks the candidate's goal relevance tags (direct match = 1.0, indirect = 0.7), then falls back to maximum cosine similarity over session goal embeddings, with a parent-chain walk-up awarding $0.7 \times \text{child\_relevance}$ for hierarchical goal structures. \textit{Global goal relevance} computes the maximum cosine similarity to any persistent goal. \textit{Recency} applies an exponential decay function:

\begin{equation}
r(t) = \exp\!\left(-\frac{\ln 2}{t_{1/2}} \cdot \Delta t\right)
\label{eq:recency}
\end{equation}

where $\Delta t$ is the time elapsed since last access (using the most recent of the last-used, updated, and created timestamps) and the half-life $t_{1/2}$ is set per profile preset, ranging from 12 hours for fast-paced worker agents (where stale context is actively harmful) to 720 hours for personal assistants that maintain long-term context, with intermediate values of 24 hours (coding), 72 hours (managerial), and 168 hours (research). \textit{Successful use prior} is the ratio of successful uses to total uses, defaulting to a configurable neutral prior for new facts. \textit{Confidence} multiplies the fact's raw confidence by a verification multiplier $m_v$ that depends on the evidence state, as described in Section~\ref{subsec:evidence}. \textit{Evidence strength} is the fraction of evidence references relative to a per-profile maximum (ranging from 2 for coding to 5 for research). \textit{Novelty} is binary: 0 if the fact's ID appears in the compaction state set (written by the compaction engine to Redis after summarization), 1 otherwise. \textit{Cost penalty} is the candidate's token count divided by the remaining budget, penalizing large entries as the budget fills.

\subsubsection{Pass 2: Interaction-Dependent Dimensions}

Two additional dimensions depend on the set of already-selected items and are therefore computed during the greedy selection loop. \textit{Redundancy penalty} computes the maximum cosine similarity between the current candidate and any already-selected item; if this exceeds a per-profile threshold (configurable per profile), a penalty is applied with weight $-0.7$. \textit{Contradiction penalty} uses a two-layer detection mechanism. The first layer checks for explicit graph edges, distinguishing between supersession edges (penalty 1.0, indicating temporal replacement) and contradiction edges (penalty 0.9, indicating logical conflict where both facts coexist). Edge direction is checked both ways to catch asymmetric relationships. The second layer identifies semantic contradictions where two highly similar facts (cosine similarity exceeding a per-profile threshold) have a large confidence gap (exceeding a per-profile threshold), applying a penalty of 0.7. The contradiction penalty carries the strongest negative weight of $-1.0$.

\subsubsection{Budget-Constrained Selection}

The final working set is assembled through greedy selection. Items flagged as mandatory (constraints, goals with blockers, procedures with required evidence) are pre-allocated unconditionally: safety requirements take precedence over budget even when this causes over-allocation, with a warning logged. The remaining candidates are sorted by Pass~1 weighted sum in descending order. For each candidate, the algorithm first checks budget fit, then recomputes the Pass~2 interaction penalties (redundancy, contradiction, and cost against the \textit{remaining} budget), recalculates the final score, and includes the candidate if the total exceeds zero. This greedy submodular approach provides a $(1-1/e) \approx 0.63$ approximation guarantee for the optimal value of the eleven-dimension objective~\cite{ref-nemhauser1978}.

\subsection{Evidence Verification Model}
\label{subsec:evidence}

The evidence system implements a four-state verification model that tracks claim trustworthiness over the lifetime of a fact. Every stored fact can be associated with typed evidence references, and the highest-priority evidence type present determines the verification state.

A newly recorded claim begins in the \textsc{Unverified} state. When any evidence is attached (for example, a reference to another supporting fact) the claim transitions to \textsc{Self-Supported}. If tool output evidence is provided (e.g., the result of an API call that confirms the claim), the state advances to \textsc{Tool-Supported}. The highest level, \textsc{Supervisor-Verified}, is reached when a human supervisor provides an explicit sign-off. Explicit rejection is possible from any state, transitioning to \textsc{Rejected} with a mandatory reason string and rejector actor ID for audit.

The verification state connects to the scoring engine through graded confidence multipliers. A supervisor-verified fact receives a multiplier of 1.0, meaning its raw confidence is fully preserved. Tool-supported facts receive 0.9, self-supported facts 0.7, facts with no associated claims receive a default of 0.8, and unverified facts receive 0.5, effectively halving their confidence in the scoring competition. This design creates a feedback loop where verified facts gain a sustained competitive advantage in working set selection, providing a structural incentive for the agent (and its operators) to invest in evidence gathering.

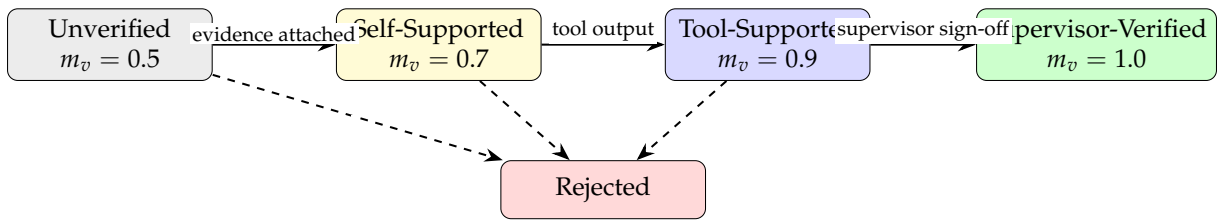
\begin{figure}[H]
\centering
\resizebox{\linewidth}{!}{%
\begin{tikzpicture}[
    state/.style={draw, rounded corners, minimum width=2.8cm, minimum height=0.8cm, align=center, font=\small},
    arrow/.style={-{Stealth[length=2.5mm]}, thick},
    lbl/.style={font=\scriptsize, fill=white, inner sep=1pt}
]
\node[state, fill=gray!15] (U) at (0,0) {Unverified\\$m_v = 0.5$};
\node[state, fill=yellow!20] (S) at (4.5,0) {Self-Supported\\$m_v = 0.7$};
\node[state, fill=blue!15] (T) at (9,0) {Tool-Supported\\$m_v = 0.9$};
\node[state, fill=green!20] (V) at (13.5,0) {Supervisor-Verified\\$m_v = 1.0$};
\node[state, fill=red!15] (R) at (6.75,-2) {Rejected};

\draw[arrow] (U) -- (S) node[midway, above, lbl] {evidence attached};
\draw[arrow] (S) -- (T) node[midway, above, lbl] {tool output};
\draw[arrow] (T) -- (V) node[midway, above, lbl] {supervisor sign-off};
\draw[arrow, dashed] (U) -- (R) node[midway, left, lbl] {};
\draw[arrow, dashed] (S) -- (R);
\draw[arrow, dashed] (T) -- (R);
\end{tikzpicture}%
}
\caption{Claim verification state machine with associated confidence multipliers $m_v$. Each transition is driven by the type of evidence attached. Dashed arrows indicate explicit rejection, which is possible from any non-terminal state.\label{fig:verification}}
\end{figure}

\subsection{Memory Management}
\label{subsec:memory_mgmt}

The memory facade provides the primary interface for storing and retrieving facts. Before persisting a new fact, the facade performs a vector similarity check against the closest existing entry. If the cosine similarity exceeds a configurable deduplication threshold, the store is skipped and a deduplication trace event is emitted.

Storage atomically creates a graph node via merge-by-ID and indexes the fact's text in a vector collection with dense embeddings. After storage, the facade creates provenance edges (authorship, actor-about, goal-serving) on a best-effort basis; edge creation failures are logged as warnings but never block the store operation. The Cognee enrichment pipeline then builds rich graph indexes for multi-hop retrieval.

The session model tracks both a stable \textit{session key} for retrieval isolation and routing, and an ephemeral \textit{session identifier} (UUID) for lifecycle detection. Session boundaries are detected through explicit hooks: session start emits a boundary trace event, and session end flushes the ingest buffer, runs the turn ingest pipeline on buffered messages, and flushes session goals to the knowledge graph. Both identifiers are stored on every fact for precise scoping.

At search time, freshness is computed as an exponential decay:

\begin{equation}
f(t) = \exp\!\left(-\frac{\ln 2}{t_{1/2}} \cdot \Delta t_{\text{hours}}\right)
\label{eq:freshness}
\end{equation}

Use counts are incremented asynchronously to avoid blocking search responses. This fire-and-forget pattern is used throughout the runtime for non-critical updates.

For GDPR compliance, the forget operation removes the node and all incident edges from the graph, performs a best-effort deletion from the vector store, and emits a GDPR deletion trace event. Following the data-minimization principle, the trace event records \textit{that} a deletion occurred without recording \textit{what} was deleted.

\subsection{Ingest Pipelines}
\label{subsec:ingest}

Three asynchronous ingest pipelines handle the transformation of raw conversational data into structured knowledge.

The turn ingest pipeline comprises ten sequential tasks. It first resolves actors mentioned in the conversation by querying the actor registry. It then loads recently extracted facts from the cache to provide context for supersession detection. The core extraction task sends the conversational turn to an LLM (with configurable temperature, token limits, and maximum facts per batch) with a structured schema requiring: category, memory class, confidence, supersession index, contradiction index (indicating logical conflict without replacement), and a goal relevance array mapping each fact to relevant goals with strength annotations (direct, indirect, or none).

Active session goals and persistent goals (with configurable caps) are injected into the extraction prompt as indexed sections, enabling goal-aware fact tagging at zero marginal LLM cost. Session goals are hint-eligible, while persistent goals are read-only context. The per-profile extraction focus further shapes the LLM's attention toward domain-relevant patterns.

When the extractor identifies supersession, the old fact's confidence is decayed by a configurable factor and a supersession graph edge is created. When it identifies contradiction without replacement, a contradiction edge is created while both facts retain their current confidence. The pipeline then classifies each fact into a memory class through the rule table described in Section~\ref{subsec:datamodel}. After batch embedding (a single call for all facts), facts are stored through the memory facade with per-profile deduplication thresholds, and the Cognee enrichment pipeline builds rich graph indexes.

Extraction also produces goal status hints, i.e., status change observations for session goals, that are dispatched to the goal hint processor even when no facts are extracted. This ensures goal status tracking is independent of fact extraction.

The artifact ingest pipeline handles tool outputs. It computes a SHA-256 hash of the tool name, sorted arguments, and output for two-level deduplication: an in-memory set catches within-process duplicates, and a persistent hash lookup catches cross-restart duplicates. Artifacts exceeding a configurable length threshold receive LLM-generated summaries; shorter artifacts are truncated directly.

The procedure ingest pipeline extracts multi-step procedures with parameterized steps and typed proof requirements. Version detection identifies updated procedures by name and scope, creating supersession edges to previous versions.

\subsection{Profile Presets and Inheritance}
\label{subsec:tiers}

Five preset profiles (coding, research, managerial, worker, and personal assistant) provide pre-tuned configurations across scoring weights, retrieval policies, isolation levels, and budget settings. Table~\ref{tab:weights} presents the full scoring weight matrix.

\begin{table}[H]
\caption{Scoring weights and configuration parameters across five profile presets.\label{tab:weights}}
\centering
\setlength{\tabcolsep}{3pt}
\begin{tabularx}{\textwidth}{lCCCCC}
\toprule
\textbf{Parameter} & \textbf{Coding} & \textbf{Research} & \textbf{Manager} & \textbf{Worker} & \textbf{Personal} \\
\midrule
turn\_relevance & 1.5 & 0.8 & 0.7 & 1.3 & 1.0 \\
session\_goal\_rel. & 1.2 & 1.0 & 1.5 & 1.4 & 0.8 \\
global\_goal\_rel. & 0.3 & 0.8 & 1.0 & 0.3 & 0.4 \\
recency & 1.2 & 0.5 & 0.6 & 1.3 & 0.9 \\
successful\_use & 0.8 & 0.6 & 0.5 & 0.7 & 0.9 \\
confidence & 0.3 & 0.8 & 0.5 & 0.4 & 0.3 \\
evidence\_strength & 0.2 & 0.9 & 0.7 & 0.3 & 0.2 \\
novelty & 0.6 & 0.7 & 0.4 & 0.5 & 0.5 \\
redundancy\_pen. & $-$0.8 & $-$0.5 & $-$0.9 & $-$0.7 & $-$0.6 \\
contradiction\_pen. & $-$1.0 & $-$1.0 & $-$1.0 & $-$1.0 & $-$1.0 \\
cost\_penalty & $-$0.4 & $-$0.2 & $-$0.5 & $-$0.4 & $-$0.3 \\
\midrule
half\_life (hours) & 24 & 168 & 72 & 12 & 720 \\
evidence\_refs\_max & 2 & 5 & 3 & 2 & 3 \\
redundancy\_thresh. & 0.85 & 0.80 & 0.90 & 0.85 & 0.85 \\
contradiction\_thr. & 0.90 & 0.85 & 0.90 & 0.90 & 0.90 \\
confidence\_gap & 0.30 & 0.25 & 0.30 & 0.30 & 0.35 \\
budget (tokens) & 8000 & 12000 & 8000 & 8000 & 8000 \\
\midrule
isolation\_level & LOOSE & NONE & LOOSE & LOOSE & STRICT \\
graph\_mode & LOCAL & GLOBAL & HYBRID & LOCAL & HYBRID \\
graph\_depth & 1 & 3 & 2 & 1 & 2 \\
\bottomrule
\end{tabularx}
\end{table}

The coding profile assigns high turn-relevance weight with a 24-hour half-life and structural-heavy retrieval (weight 0.5), favoring fresh, topically aligned code decisions. The research profile emphasizes confidence and evidence strength with a 168-hour half-life, deep graph traversal (GLOBAL mode, depth 3), and vector-heavy retrieval (weight 0.5). The managerial profile prioritizes session goal relevance (1.5) with graph-heavy retrieval oriented toward relationship discovery. The worker profile mirrors coding's structural emphasis with a shorter 12-hour half-life for ephemeral task contexts. The personal assistant profile uses strict isolation for privacy and a 720-hour half-life for long-term memory retention.

Profiles support a four-level inheritance chain: base $\to$ profile preset $\to$ organization override $\to$ resolved policy. The resolved policy is flattened at bootstrap and remains immutable for the session lifetime. Organization overrides are stored as configuration entities in the knowledge graph, keyed by organization and profile identifiers. This allows a financial services organization to override evidence strength from 0.2 to 0.8 and add custom guard rules without forking the base profile. Auto-recall uses a separate retrieval policy with a minimum similarity floor (0.3) and reduced top-$k$ (10), distinct from the explicit search policy.

\subsubsection{Authority-Based Access Control}

The system implements a numeric authority model, analogous to Unix privilege levels, that governs which actors can perform which operations. Every actor carries an integer \texttt{authority\_level} field. Eleven configurable authority rules, stored in an SQLite-backed rule store and customizable per deployment, define the minimum authority level and optional constraints for each administrative action.

Four authority tiers partition the permission space. \textit{Regular actors} (level~0--49) can create goals and session data within their own scope. \textit{Team leads} (50--69) can additionally create team-scoped goals and manage team membership. \textit{Organization administrators} (70--89) can create organization-scoped goals, register actors, set profile overrides, and merge duplicate actors. \textit{System administrators} (90+) can create global goals, manage cross-organization configuration, and override any lower-level constraint.

Each authority rule specifies a \texttt{matching\_exempt\_level}: actors at or above this level bypass the rule's organizational or team membership checks, analogous to the Unix root user bypassing file permission checks. For example, team membership management requires authority level~50 and matching team membership, but actors at level~70 (the exempt level) can manage any team in their organization without being a member. This creates a layered delegation model where higher authority implicitly encompasses lower-level permissions.

Guard verdicts of \textit{require\_approval} invoke the authority model: only supervisors with sufficient authority level for the specific violated rule can provide the sign-off. The approval is recorded as supervisor-verified evidence, feeding back into the verification model (Section~\ref{subsec:evidence}). A cross-session pending approval index enables supervisors to review all outstanding approvals across agents from a unified queue.

Organizations and teams are modeled as first-class graph entities that span gateway boundaries, distinguishing business-level identity (which organization does this actor belong to?) from infrastructure-level identity (which gateway is this request routed through?). Actors carry platform-qualified handles (\texttt{email:alice@acme.com}, \texttt{telegram:alice\_tg}) that enable cross-platform identity resolution.

\subsection{Session Goal Management}
\label{subsec:goals}

The goal manager maintains a two-bucket model. \textbf{Bucket~1} holds session-scoped goals in the cache with a configurable TTL, supporting real-time CRUD through five TypeScript tools. Only ACTIVE and PROPOSED goals are included in working set candidates. \textbf{Bucket~2} holds persistent goals in the Neo4j knowledge graph with scope annotations (GLOBAL, ORGANIZATION, TEAM, ACTOR) linked to owning actors through ownership edges. The persistent goal query uses graph pattern matching with optional actor filtering: goals without owners are visible to all actors, while owned goals require an actor identity match. Persistent goals support blocker management and sub-goal creation through the admin API, enabling hierarchical goal decomposition at the organizational level.

Goals form parent--child hierarchies through parent--child edges. The scoring engine walks the parent chain to award relevance bonuses: direct matches receive 1.0, indirect matches 0.7, and parent goals receive $0.7 \times \text{child\_relevance}$. A goal refinement pipeline processes six hint types in two tiers: Tier~1 hints (completed, abandoned, blocked, progressed) are applied directly to Redis at zero LLM cost, while Tier~2 hints (refined, new subgoal) invoke the LLM asynchronously. When a sub-goal completes, the parent goal's confidence is updated to the ratio of completed siblings. Sub-goal creation includes Jaccard deduplication against sibling titles and a per-session cap.

At session end, session goals are flushed to the knowledge graph, with status annotations (completed criteria, abandonment reasons, or active blockers) indexed for future retrieval, and parent--child and ownership edges created. A persistent audit store records all session goal lifecycle events (created, updated, completed, abandoned, blocked, progressed, refined, subgoal created, flushed).

\subsection{Procedure Lifecycle Tracking}
\label{subsec:procedure_lifecycle}

Procedures represent standard operating procedures (SOPs): stored, versioned workflow templates that prescribe how an agent should accomplish a specific task. Unlike ad-hoc instructions, procedures are persistent graph entities with defined steps, activation conditions, proof requirements, guard bindings, and optional organization and team scoping that survive across sessions.

\subsubsection{Procedure Structure}

A procedure definition contains an ordered list of steps, each with a natural-language instruction and optional proof requirements. Five proof types are supported: \textit{diff hash} (cryptographic proof of a code change), \textit{chunk reference} (pointer to a specific document passage), \textit{receipt} (external system confirmation such as a deployment log), \textit{version record} (before/after state proof), and \textit{supervisor sign-off} (explicit human approval). Each proof requirement specifies whether it is mandatory or advisory, enabling flexible enforcement. Procedure-level guard bindings declare safety constraints that become active when the procedure is in use; for example, a ``Deploy to Production'' procedure may bind a ``no unreviewed deploys'' constraint and a ``no deploy during freeze'' constraint. When a procedure activates, its bindings are loaded into the guard engine's static rule registry, where each binding string becomes both a keyword pattern rule and a semantic exemplar for Layer~2 matching. The guard surface thus \textit{grows} dynamically as procedures activate: a ``Deploy to Production'' procedure adds deploy-safety constraints, while a concurrent ``Handle Customer PII'' procedure adds data-handling constraints, and both rule sets apply simultaneously. When a procedure completes, its bindings are unloaded, returning the guard surface to its baseline state. This dynamic rule composition means that the set of active safety constraints is contextual --- driven by which workflows are currently executing --- rather than a static rulebook.

\subsubsection{Activation and Execution}

Procedures support six activation modes: \textit{manual} (agent or user explicitly starts), \textit{actor default} (auto-activated for specific actor roles), \textit{trigger word} (activated when a keyword appears in conversation), \textit{task classifier} (LLM determines the procedure matches the current task), \textit{goal-bound} (activated when an associated goal becomes active), and \textit{supervisor-forced} (a supervisor mandates the procedure). Upon activation, three things happen simultaneously: (1)~the procedure's steps are injected into the working set as mandatory context so the agent sees what to do next, (2)~any guard bindings are loaded into the guard engine's static rule registry, adding safety constraints specific to this workflow, and (3)~an execution record is created to track step completion, proof submission, and elapsed time.

As the agent works through the procedure, it reports step completions with optional proof artifacts. The engine tracks which steps are done, which proofs have been submitted, and which remain outstanding. Steps can be completed in any order unless the procedure specifies sequential dependencies.

\subsubsection{Completion Verification}

A procedure cannot be marked complete until a completion gate verifies that all required proofs have been satisfied. For each mandatory proof requirement on each step, the gate queries the evidence engine to check whether a matching claim exists with the correct evidence type and a verification status of at least \textsc{Self-Supported}. Missing proofs are surfaced to the agent as context (``Step~3 requires a deployment receipt, not yet provided'') and, if the procedure binds a guard constraint, may trigger a \textit{require\_evidence} guard verdict that blocks further progress until proof is submitted. When the procedure completes, its guard bindings are unloaded from the static rule registry, and a completion event with all proof references is recorded in the persistent audit store.

\subsubsection{Working Set Integration}

Procedure candidates in the working set are filtered by an enable gate, optional relevance filtering with a cosine similarity threshold, a top-$k$ cap, and an override flag that always includes procedures requiring proof regardless of relevance score. Procedures with outstanding proof requirements receive the mandatory-injection flag, ensuring they appear in the context even when budget is tight. Version detection identifies updated procedures by name and scope, creating supersession edges to previous versions. A persistent audit store records all procedure lifecycle events (qualified, activated, step completed, proof submitted, completed, abandoned). Four TypeScript tools expose procedure operations to the agent.

\subsection{Embedding Cache}
\label{subsec:embedding_cache}

To reduce embedding API costs in production, the runtime wraps the embedding service with a cache backed by Redis. Cache keys are derived from truncated SHA-256 hashes of the input text. Batch operations use pipelined reads and writes (one round-trip each); only cache misses are forwarded to the embedding API. The cache is configurable with a per-deployment TTL and degrades gracefully on cache failure by falling through to the uncached embedding service.

\subsection{Context Lifecycle}
\label{subsec:context_lifecycle}

The context lifecycle orchestrates five stages per conversational turn, managing the transition from raw messages to assembled, guarded context.

\textbf{Bootstrap.} On session start, the lifecycle resolves the profile policy (immutable for the session lifetime), registers the agent actor if not already present, loads session goals from Redis, and creates a session context record in the cache containing the resolved profile, turn counter, compact state IDs, procedure execution references, and goal injection history. The session context persists across lifecycle calls and is deleted at session end.

\textbf{Ingest.} The turn ingest pipeline (Section~\ref{subsec:ingest}) processes conversational messages, while the artifact ingest pipeline handles tool outputs. Goal status hints are dispatched to the hint processor. Auto-capture of tool outputs follows a configurable policy per profile: coding profiles capture all tool outputs; research profiles capture only search and analysis outputs.

\textbf{Assemble.} The working set manager (Section~\ref{subsec:scoring}) produces a scored snapshot. The guard engine (Section~\ref{subsec:guards}) performs a preflight check, potentially adding constraint reinjections as mandatory items. The context assembler then distributes items across four blocks on two prompt surfaces. \textit{Surface~A} (system prompt addition, highest attention): Block~1 contains constraints, guard reinjections, and proof-required procedure instructions. \textit{Surface~B}: Block~2 (prepended system context) contains goals with blockers highlighted; Block~3 (prepended context) contains the scored working set ordered by memory class (POLICY $\to$ PROCEDURAL $\to$ SEMANTIC $\to$ EPISODIC) then by score descending; Block~4 (appended system context) contains evidence references. A per-profile assembly placement policy controls which blocks are active: research profiles disable constraints and guards for uninterrupted work, while managerial profiles always show goals with blocker highlighting.

Tool output replacement reduces token usage: old tool outputs (exceeding 100 tokens and not among the last $N$ messages) are replaced with approximately 40-token placeholders containing the tool name, a brief summary, and a search reference, with content deduplicated by SHA-256. A session artifact store tracks injection and search counts for consolidation-time promotion decisions.

Conversation deduplication, when enabled per profile, removes tool-role messages from the conversation that are already covered by Block~3 working set items (Jaccard overlap threshold 0.7), preventing double-counting of the token budget.

\textbf{Compact.} Two-stage compaction reduces conversation history token usage. Stage~1 (zero LLM cost) classifies each message as PRESERVE (decisions, goals, evidence, last agent message), COMPRESS (verbose outputs exceeding 500 tokens, same-topic sequences), or DROP (duplicates, filler). Stage~2 (one LLM call) summarizes the COMPRESS bucket into one to three paragraphs. Cadence thresholds vary by profile aggressiveness: aggressive profiles trigger at $1.3\times$ the target token count, balanced at $2\times$, and minimal at $4\times$. After compaction, the IDs of compacted items are written to a Redis set, which the scoring engine's novelty dimension reads: compacted items receive a novelty score of 0, suppressing their re-injection into the working set.

\textbf{After-Turn.} The lifecycle detects which injected facts were successfully used through six signals. \textit{S1~Direct Quote} (weight 0.95): substring match of the fact text in the agent's response. \textit{S2~Tool Correlation} (0.70): the response invokes a tool whose alias appears in the fact. \textit{S3~Goal Mention} (0.50): a fact tagged with a goal-serving relationship references a goal discussed in the response. \textit{S4~Emergent Behavior} (0.40): LLM-based correlation between the fact and the response. \textit{S5~Continuation} (0.35): high semantic similarity between the fact and the response. \textit{S6~Ignored Penalty}: facts injected for three or more turns with zero signals suppress their use prior. A running Jaccard coefficient exceeding 0.3 triggers a successful-use increment. This signal feeds the scoring engine's successful-use-prior dimension, creating a closed-loop feedback system. For deployments requiring higher-fidelity signals, an optional LLM-based successful-use reasoning task can replace the heuristic signals with explicit correlation analysis between injected facts and agent responses. A companion blocker extraction task identifies goal-blocking conditions mentioned in conversation, dispatching them as session goal blockers at zero manual annotation cost. Both tasks are configurable per profile and disabled by default to minimize LLM cost.

Smart goal injection applies a cadence policy: goals are re-injected only when at least one of five conditions holds: first turn, status changed, blocker changed, reminder interval reached (every 5 turns), or agent engagement detected (the agent mentioned the goal). This avoids injecting unchanged goals on routine turns.

Subagent support is provided through two hooks. The spawn hook filters the parent working set by child goal relevance with a $2\times$ budget allocation, stores the parent--child mapping in the cache, and creates a subagent-inherit isolation scope for the child session. The end hook merges child facts into the parent session, updates parent goals based on child outcomes, and cleans up the cache mapping.

\subsection{Guard Pipeline}
\label{subsec:guards}

The guard engine implements a six-layer cheap-first escalation pipeline that enforces safety constraints on every agent action during context assembly.

\textbf{Layer~0: Autonomy Domain Classification.} Every agent action is first classified into one of ten autonomy domains: FINANCIAL, DATA\_ACCESS, COMMUNICATION, CODE\_CHANGE, SCOPE\_CHANGE, RESOURCE, INFORMATION\_SHARE, DELEGATION, RECORD\_MUTATION, and UNCATEGORIZED. Each domain is assigned a per-profile autonomy level drawn from a four-point escalation scale: \textit{AUTONOMOUS} (the agent may proceed without notification), \textit{INFORM} (the agent proceeds but the action is logged for later review), \textit{APPROVE\_FIRST} (execution is suspended until a human supervisor explicitly approves), and \textit{HARD\_STOP} (the action is unconditionally blocked regardless of any other signal). The autonomy floor for a given action is the level assigned to its domain in the active profile. This floor is then composed with the safety verdict from layers 1--4 via the rule $\text{final} = \max(\text{autonomy\_floor}, \text{safety\_result})$, ensuring that safety checks can only escalate the required oversight level, never relax it. This composition means that even if layers 1--4 return \textit{pass}, an action in a domain set to APPROVE\_FIRST will still require human sign-off.

The autonomy matrix is fully configurable per profile, enabling fine-grained control over agent authority. For example, the coding profile sets FINANCIAL to HARD\_STOP (agents must never initiate financial transactions) and CODE\_CHANGE to AUTONOMOUS for development environments but APPROVE\_FIRST for production, while the managerial profile sets SCOPE\_CHANGE to AUTONOMOUS (managers can restructure goals freely) but CODE\_CHANGE to HARD\_STOP (managers should not modify code). Organizations can override individual domain levels through the profile inheritance chain, allowing a financial services organization to tighten DATA\_ACCESS to APPROVE\_FIRST across all profiles without modifying the base configuration.

\textbf{Layer~1: Static Rule Matching.} A four-source rule registry merges built-in rules, profile-specific rules, procedure-bound rules (loaded dynamically from active procedures' guard bindings on activation), and operator-defined custom rules (created through the admin API with per-gateway and per-organization scoping). Rules match via keyword, phrase, regex, or tool target patterns. This layer catches obvious violations (e.g., ``delete production database'') with zero LLM cost.

\textbf{Layer~2: Semantic Matching.} BM25 scoring (block threshold 8.0, warn threshold 5.0) and cosine similarity (threshold 0.80) compare the agent's intended action against known-bad exemplars in the guard rule set. A per-profile strictness modifier adjusts thresholds: loose profiles multiply by 1.5 (harder to trigger), strict profiles by 0.7 (easier to trigger).

\textbf{Layer~3: Structural Validation.} Graph queries check for approval tokens (has a supervisor approved this action?), evidence existence (has the agent provided proof for its claim?), and confirmation flags. Disabled under loose strictness.

\textbf{Layer~4: Forced Constraint Reinjection.} Unlike the preceding layers, which determine whether an action is safe, this layer always executes regardless of prior verdicts. It collects all triggered constraints (from layers 0--3 or from active procedure bindings) and formats them as mandatory working set items for injection into Block~1 of the context assembly. This ensures the agent sees the full set of applicable constraints in its system prompt, even when the verdict is \textit{pass} or \textit{warn}.

\textbf{Layer~5: LLM Escalation.} A narrow safety-focused prompt is sent to the LLM for ambiguous cases that pass all cheaper layers. This stage fires only under strict profiles or when explicitly enabled. Under medium strictness, it fires only when all prior layers return non-pass results.

After evaluation, the guard engine produces one of five result types: \textit{pass} (safe to proceed), \textit{warn} (suspicious but allowed, logged), \textit{block} (action prevented), \textit{require\_approval} (held for human sign-off), or \textit{require\_evidence} (agent must provide proof first). Constraints that trigger blocks or warnings are reinjected into the working set as mandatory items, ensuring the agent sees the violated rule in its next context assembly.

The guard pipeline serves as a human-in-the-loop (HITL) middleware: when an action receives a \textit{require\_approval} or \textit{require\_evidence} verdict, execution is suspended and the decision is surfaced to a human supervisor through the plugin interface. The supervisor can approve, reject, or escalate, and the decision is recorded as a typed evidence reference (specifically, a supervisor sign-off) that feeds back into the verification model. This design treats human oversight not as an external bolt-on but as an integral part of the cognitive loop, where supervisor decisions become first-class evidence that strengthens the trustworthiness of the knowledge base over time.

To illustrate the interplay between procedures, guards, autonomy, and HITL, consider a concrete scenario. A coding agent activates a ``Deploy to Production'' procedure, which binds two guard constraints (``no unreviewed deploys'' and ``no deploy during freeze'') and requires a deployment receipt as proof for the deploy step. The agent first runs the test suite and submits a chunk reference as proof. It then attempts to deploy. Layer~0 classifies this as a CODE\_CHANGE action; the coding profile sets this domain to APPROVE\_FIRST in production. Even though layers 1--4 return \textit{pass} (no static rule or semantic match triggered), the autonomy floor escalates the verdict to \textit{require\_approval}. The action is suspended and surfaced to a human tech lead, who reviews the test results and approves. The approval is recorded as a supervisor sign-off evidence reference, the fact confidence for the deployment claim rises (the supervisor-verified multiplier of 1.0 replaces the previous 0.5 for unverified), and the deploy proceeds. The agent then submits the deployment receipt, the completion gate verifies all proofs are satisfied, and the procedure is marked complete. Had the agent attempted to deploy during a code freeze, Layer~1 would have matched the ``no deploy during freeze'' static rule, producing a \textit{block} verdict that no amount of supervisor approval could override; the constraint would be reinjected as mandatory context explaining why the action was blocked.

A near-miss escalation mechanism tracks guard results in a per-session list. When three near-misses (warn results) occur within five turns, the strictness level is automatically tightened (medium $\to$ strict, LLM layer enabled), preventing boundary-probing behavior from gradually circumventing the guard system.

The six-layer pipeline described above produces \textit{advisory} enforcement: constraints are injected into the agent's system prompt (Block~1, Surface~A), where the LLM sees them and is expected to comply. Advisory enforcement is effective in approximately 95\% of cases and has the advantage of contextual understanding --- the agent knows \textit{why} an action is restricted and can propose alternatives constructively. However, advisory enforcement cannot physically prevent a tool call that violates a constraint. Section~\ref{subsec:firewall} describes the AI Firewall, which adds physical enforcement through tool-call interception, closing this gap.

The interplay between procedures, guards, and evidence forms a triangle of accountability. When a procedure activates, its guard bindings load as static rules (connection~1). When a guard detects a violation that requires evidence, the constraint is reinjected and the agent records a claim against it (connection~2). When a procedure completion check runs, the verification engine validates that all proof requirements have corresponding verified claims (connection~3). Table~\ref{tab:triangle} contrasts the three subsystems across eight dimensions.

\begin{table}[H]
\caption{Comparison of procedures, guards, and evidence across eight dimensions.\label{tab:triangle}}
\centering
\begin{tabularx}{\textwidth}{lCCC}
\toprule
\textbf{Dimension} & \textbf{Procedures} & \textbf{Guards} & \textbf{Evidence} \\
\midrule
Nature & Prescriptive & Restrictive & Accountable \\
Timing & When activated & Every action & After action \\
Question & How to do it? & Is it safe? & Can you prove it? \\
Scope & Per-workflow & Per-session & Per-claim \\
Failure mode & Incomplete step & Blocked action & Unverified claim \\
Storage & Cognee graph & Redis + graph & Cognee graph \\
Profile-driven & Role variants & Strictness levels & Sampling rates \\
Analogy & Recipe & Health code & Inspection \\
\bottomrule
\end{tabularx}
\end{table}

\subsection{Consolidation Engine}
\label{subsec:consolidation}

The consolidation engine runs during configurable idle periods to perform nine stages of memory maintenance, analogous to the role of sleep in human memory processing~\cite{ref-anderson2004}: strengthening useful synaptic connections (fact reinforcement), weakening irrelevant ones (confidence decay), promoting recurring episodic memories to semantic storage (memory consolidation), and extracting reusable skills from repeated patterns (procedure refinement).

\textbf{Stage~1: Cluster Near-Duplicates.} Embedding cosine similarity at a threshold of 0.92 identifies clusters of near-identical facts. No LLM calls are made; this is pure vector math.

\textbf{Stage~2: Canonicalize.} For each cluster, a canonical representative is selected by majority voting over text variants. LLM resolution is used only for ambiguous clusters (fewer than 5\% of cases), keeping LLM costs minimal.

\textbf{Stage~3: Strengthen Useful Facts.} Facts that have been recalled and used receive a confidence boost:

\begin{equation}
c'(f) = \min\!\left(c(f) + \frac{s(f)}{u(f)} \cdot 0.3,\; 1.0\right)
\label{eq:strengthen}
\end{equation}

where $s(f)$ is the successful use count and $u(f)$ is the total use count.

\textbf{Stage~4: Decay Unused Facts.} Two decay modes operate on different timescales. Facts that have been recalled but never used decay as $c' = c \cdot 0.9^{t_r}$ where $t_r$ is the number of recall turns. Facts that have never been recalled decay as $c' = c \cdot 0.95^{t_d}$ where $t_d$ is days since creation. Decay rates vary by scope: SESSION-scoped facts decay hourly, ACTOR-scoped daily, TEAM and ORGANIZATION-scoped weekly, and GLOBAL-scoped monthly. Importantly, decay is computed based on the last-used timestamp rather than the creation timestamp: a long-lived fact that is continuously useful retains its confidence through the ``touch'' mechanism of working set appearance. Facts whose confidence falls below 0.1 are archived.

\textbf{Stage~5: Prune Auto-Recall.} Facts that have been auto-recalled five or more times with zero successful uses are blacklisted from future auto-recall, preventing the system from repeatedly injecting information the agent never acts on.

\textbf{Stage~6: Promote.} Recurring episodic facts (appearing in three or more sessions) are promoted to semantic class. Critically, class and scope must co-promote: a SEMANTIC fact at SESSION scope is ``durable but invisible'' (it will not decay but is inaccessible under STRICT isolation). A decision matrix considers recurrence, goal linkage, and use count to determine whether to promote both class and scope, class only, or neither.

\textbf{Stage~7: Refine Procedures.} Step patterns recurring across three or more sessions are identified and proposed as new procedure definitions. This is one of only two stages requiring LLM calls, invoked once per detected pattern.

\textbf{Stage~8: Identify Verification Gaps.} Claims are compared against their procedure's proof requirements. Missing evidence is flagged for the agent or queued for supervisor sampling at a configurable rate.

\textbf{Stage~9: Recompute Salience.} An exponential moving average (EMA) adjusts fact salience priors based on cumulative use data, capped at $\pm 5\%$ per consolidation cycle to prevent oscillation. Weight adjustments are scoped per (profile, org, gateway), ensuring no cross-gateway tuning interference.

Consolidation execution is gateway-scoped: each run processes facts for a single gateway identifier, ensuring strict isolation between deployments. A Redis distributed lock with a one-hour timeout prevents concurrent consolidation runs on the same gateway. The scoring weight adjustments from Stage~9 are persisted in an SQLite-backed tuning delta store, keyed by profile, organization, and gateway, enabling the weight resolution chain (base preset $\to$ profile $\to$ organization override $\to$ tuning delta) to incorporate learned adjustments across runs. A configurable audit store with retention-based rotation records consolidation reports for operational review.

In total, seven of nine stages require zero LLM calls. Stage~2 uses the LLM for fewer than 5\% of clusters, and Stage~7 uses one call per detected pattern, resulting in 2--5 LLM calls per consolidation run. Artifacts are promoted to persistent storage when their search count exceeds zero or their injection count reaches three; others expire with their cache time-to-live.

\subsection{AI Firewall}
\label{subsec:firewall}

The six-layer guard pipeline (Section~\ref{subsec:guards}) produces advisory verdicts: constraints are injected into the agent's system prompt, and the language model is expected to comply. While advisory enforcement is effective in approximately 95\% of cases and provides contextual understanding (the agent knows \textit{why} an action is restricted), the language model remains the final arbiter of whether to follow the injected constraint. For high-stakes operations such as financial transactions, data deletion, or production deployments, this compliance gap is unacceptable.

\subsubsection{Dual-Enforcement Architecture}

The AI Firewall closes the advisory compliance gap through physical enforcement at three layers, operating in parallel with the advisory path.

\textit{Layer~A (agent hook):} The TypeScript ContextEngine plugin intercepts every tool invocation via the host platform's \texttt{onToolCall} lifecycle hook before the tool executes. A fast-path cache lookup returns cached guard verdicts in under 5\,ms for previously evaluated tool--argument combinations; cache misses invoke the full six-layer guard pipeline synchronously. If the verdict is \textit{block} or \textit{require\_approval} (without a pending approval), the tool call is rejected at the runtime level --- the tool never executes, and the agent receives an error message explaining the reason. This is the primary enforcement point.

\textit{Layer~B (proxy guardrails):} A standalone LiteLLM~\cite{ref-litellm} guardrails server sits between the agent and the LLM API, scanning both prompts (pre-call) and completions (post-call) through configurable guard chains. This catches threats at the API boundary: prompt manipulation, completion-level data exfiltration, and tool-call requests embedded in completions that bypass Layer~A.

\textit{Layer~C (SDK hooks):} For applications using the LiteLLM Python SDK directly, in-process guardrails hooks apply the same guard logic without an HTTP round-trip.

The advisory and physical paths are complementary. Advisory enforcement (Section~\ref{subsec:guards}) runs during context assembly, proactively informing the agent of constraints so it can plan around them constructively. Physical enforcement runs at tool invocation and API boundary time, reactively catching violations that slip through advisory compliance. Together they form a belt-and-suspenders model: advisory prevents most violations through informed compliance; physical catches the remainder through hard enforcement.

\subsubsection{Cost-Tiered Safety Scanning}

Orthogonal to guard enforcement, the AI Firewall adds input/output safety scanning to detect threats that rule-based and semantic matching cannot catch. A safety guard engine implements six tiers invoked in cost order, with early short-circuiting on high-confidence detection:

\textit{Tier~0 (pattern detection):} Approximately 200 regex patterns derived from OWASP LLM security guidelines~\cite{ref-owasp-llm} detect common prompt injection templates, known jailbreak patterns, and data exfiltration attempts. Zero LLM cost, sub-millisecond latency.

\textit{Tier~1 (heuristic scoring):} Statistical analysis of token entropy, special character density, instruction-like phrase frequency, and encoding anomalies flags suspicious inputs without model invocation.

\textit{Tier~2 (classification models):} Fine-tuned models from LLM~Guard~\cite{ref-llmguard} perform prompt injection detection, PII identification, toxicity scoring, and relevance assessment. Moderate cost, approximately 100\,ms latency.

\textit{Tier~3 (declarative validators):} Guardrails~AI~\cite{ref-guardrailsai} validators enforce structural and content constraints on LLM outputs with automatic correction and retry.

\textit{Tier~4 (dialog rails):} NeMo Guardrails~\cite{ref-nemo} rails detect topic drift, refusal bypass attempts, and multi-turn manipulation patterns at the conversation level.

\textit{Tier~5 (canary tokens):} Inspired by Rebuff~\cite{ref-rebuff}, traceable tokens are injected into stored content to detect unauthorized exfiltration across system boundaries.

Per-profile safety policies control which tiers are active and their sensitivity thresholds. Libraries that are not installed degrade gracefully: the engine skips unavailable tiers and logs a warning, ensuring that the safety pipeline operates at the best available fidelity without hard dependencies.

\subsubsection{Knowledge Base Contamination Prevention}

Safety scanning is integrated with the ingest pipeline, not only with the guard pipeline. When a scanner flags a tool output as containing prompt injection or adversarial content, the ingest pipeline skips fact extraction for that output and records a safety scan trace event with the detection tier and confidence score. This prevents adversarial content from being stored as a trusted fact in the knowledge graph, where it could persist across sessions and influence future agent behavior through the retrieval pipeline --- a particularly dangerous vector unique to persistent memory systems.

%%%%%%%%%%%%%%%%%%%%%%%%%%%%%%%%%%%%%%%%%%
\section{Validation Results}
\label{sec:results}

We evaluate ElephantBroker through architectural validation using a comprehensive test suite organized across unit, integration, and end-to-end levels. Because the system is a runtime infrastructure component, its correctness properties are best verified through systematic testing rather than downstream task benchmarks, analogous to how a database engine is validated through correctness tests before measuring application-level performance.

\subsection{Scoring Engine Correctness}

The scoring engine is the most algorithmically complex component. Dimension independence tests confirm that modifying one scoring dimension does not affect others. Recency decay correctness is verified at boundary conditions: $r(0) = 1.0$, $r(t_{1/2}) = 0.5$, $r(2 t_{1/2}) = 0.25$ for each profile's configured half-life, matching the expected exponential decay from Equation~\ref{eq:recency}. Budget enforcement tests use randomized candidate sets to verify that the greedy selector never exceeds the configured token budget. Redundancy detection tests confirm that candidates exceeding the per-profile similarity threshold receive the penalty, and contradiction detection tests verify both graph-edge-based and semantic-similarity-based conflict identification across all per-profile threshold combinations.

Critically, verification multiplier tests confirm that facts with higher verification states consistently outrank facts of equal raw confidence but lower verification. This property ensures that the evidence model's trustworthiness signal propagates correctly through the scoring pipeline.

\subsection{Retrieval Pipeline Correctness}

Integration tests verify that all five retrieval sources return correctly typed candidates when queried against populated graph and vector stores. Source weight multiplication produces expected merged rankings in controlled scenarios. Isolation policy tests confirm that strict mode returns only scoped results (with keyword source disabled), loose mode applies post-filtering, and none mode returns all matches. The four-stage reranking pipeline is tested for ordering invariants and graceful degradation when the cross-encoder endpoint is unavailable.

\subsection{Evidence Model Correctness}

The verification state machine is tested for correct transitions across all evidence type combinations. Determinism tests verify that the same evidence set always produces the same verification state, regardless of the order in which evidence is attached. Integration tests confirm that confidence multipliers correctly modulate scoring, and audit trail tests verify that all state transitions are recorded in the trace ledger.

\subsection{Classification, Isolation, and Plugin Compatibility}

Memory class classification is tested against the full rule table: each of the twelve built-in categories maps to the expected memory class, custom categories trigger the LLM fallback path, and the ambiguity resolution produces deterministic results. Isolation level tests verify that STRICT mode disables the keyword retrieval source and uses direct vector payload filters, that LOOSE mode post-filters by the configured scope, and that scope visibility respects the eight-level hierarchy across all five profile defaults. End-to-end tests verify that the TypeScript plugins correctly invoke all twenty-four tools and that auto-recall and auto-capture hooks function within the OpenClaw lifecycle. W3C trace propagation is verified across the TypeScript--Python boundary.

\subsection{Guard Enforcement and Safety Scanning}

Tool-call interception tests verify that blocked verdicts physically prevent tool execution and return appropriate error responses. Cache hit tests confirm that repeated tool--argument combinations return cached verdicts within the 5\,ms latency target. Safety scanning tier tests verify detection of known prompt injection patterns (Tier~0), heuristic anomaly flagging (Tier~1), and graceful degradation when optional scanning libraries are not installed. Knowledge base contamination tests confirm that flagged tool outputs are excluded from fact extraction while clean outputs proceed normally.

%%%%%%%%%%%%%%%%%%%%%%%%%%%%%%%%%%%%%%%%%%
\section{Discussion}
\label{sec:discussion}

\subsection{Trustworthiness Through Architecture}

ElephantBroker addresses several dimensions of trustworthy AI through architectural design rather than post-hoc mitigation. The evidence verification model provides a persistent mechanism for factuality: stored facts are not treated as equally reliable but are graded by their evidential support, and this grading directly influences context assembly priority. The two-layer contradiction detection system (graph-edge-based for explicit conflicts and semantic-similarity-based for implicit ones) helps identify inconsistencies in the knowledge base. The audit trail, with 47 event types spanning the full lifecycle from fact extraction through scoring to GDPR deletion, enables post-hoc verification of agent decisions. The goal-aware extraction pipeline, which injects active goals into fact extraction prompts, ensures that goal relevance is captured at ingestion time rather than computed solely at retrieval time, reducing the gap between what the agent knows and what it can surface when needed.

The AI Firewall (Section~\ref{subsec:firewall}) extends trustworthiness from retrieval-time guarantees to ingest-time protection. By scanning tool outputs for prompt injection and adversarial content before fact extraction, the system prevents knowledge base contamination --- a threat vector unique to persistent memory systems where a single malicious injection can influence agent behavior across all future sessions through the retrieval pipeline. The dual-enforcement architecture ensures that guard verdicts are not merely advisory suggestions but physically enforced constraints, with tool-call interception providing a hard barrier that the language model cannot bypass.

\subsection{Cost Optimization as a Design Principle}

Throughout the system, cost minimization is a first-class design concern rather than an afterthought. Goal piggybacking in the extraction prompt provides goal-aware fact tagging at zero marginal LLM cost. Batch embedding uses a single API call per $N$ facts. The embedding cache reduces repeated calls to two round-trips per batch via pipelined reads and writes. Four of six goal refinement hint types require zero LLM calls (Tier~1 direct updates). Seven of nine consolidation stages are pure computation. Guard layers 0--3 (autonomy, static, semantic, structural) require no LLM invocation. Two-stage compaction applies the LLM only to the COMPRESS bucket, not the entire conversation. SHA-256 artifact deduplication prevents redundant storage and summarization. Smart goal injection cadence suppresses injection on routine turns. Fire-and-forget use-count updates avoid blocking critical paths.

\subsection{Enterprise Compliance Triangle}

In regulated environments, the procedure--guard--evidence triangle maps directly to compliance frameworks. Procedures correspond to standard operating procedures required by SOX and HIPAA: auditable, versioned, with proof-of-completion gates. Guards correspond to safety enforcement required by the EU AI Act~\cite{ref-euaiact}: real-time action validation with escalation policies. Evidence corresponds to audit trails: every claim can be traced to its proof, every guard decision to its rule, and every procedure completion to its verified requirements. The interconnection between these three subsystems (procedure activation loads guard constraints, guard violations demand evidence, completion verification checks evidence) ensures that compliance is enforced structurally rather than relying on agent good behavior. Profile inheritance with organization overrides enables enterprise customization (e.g., a financial services organization overriding evidence thresholds) without forking the base system.

The AI Firewall adds an enforcement dimension to the compliance triangle: procedures define what must be done, guards determine what is safe, evidence proves what was done, and the firewall ensures that unsafe actions are physically prevented rather than merely discouraged. The authority model provides the authorization layer: each vertex of the triangle (procedure creation, guard rule management, evidence sign-off) is gated by authority level thresholds, ensuring that only actors with sufficient privilege can modify the compliance surface. Red-line bindings on procedures serve as the formal bridge between prescriptive workflows and restrictive guards: activating a regulated procedure automatically activates its safety constraints, making the compliance surface contextual rather than static.

\subsection{Scenario-Based Validation and Trace Infrastructure}

The system includes a scenario-based validation framework with a composite reward function for systematic correctness assessment. Seven scenarios exercise the full cognitive loop: basic memory store-and-recall, multi-turn memory consistency, goal-driven context assembly, context lifecycle transitions, subagent delegation with knowledge inheritance, procedure execution with proof gating, and guard pipeline activation with escalation. Each scenario produces a reward score composed of step completion (60\% weight), trace assertion satisfaction (40\%), minus an error penalty. A five-component aggregate reward function weights unit tests (0.30), integration tests (0.20), scenario outcomes (0.25), trace event health (0.15), and health check responses (0.10).

An enhanced trace API supports diagnostic workflows through full-predicate trace queries (filtering by session, gateway, event types, actors, and time ranges), turn-grouped session timelines, and aggregate session summaries with 18 computed fields covering event counts, duration statistics, and type distributions. Guard events, scoring decisions, and consolidation reports flow through a TraceLedger--OTEL bridge to ClickHouse for durable cross-session analytics, enabling operators to query guard outcomes and retrieval patterns across arbitrary time windows.

\subsection{Comparison with Existing Systems}

Compared with MemGPT~\cite{ref-packer2024}, ElephantBroker provides richer knowledge representation through graph-structured memory with typed relationships and evidence tracking, rather than treating memory as opaque text blocks. Compared with standard RAG pipelines~\cite{ref-lewis2020}, it maintains persistent, evolving memory that grows with each conversational turn. Compared with GraphRAG~\cite{ref-edge2024}, it constructs its knowledge graph incrementally from live interactions. Compared with Zep~\cite{ref-zep2025}, which organizes memory into episodic, semantic, and community tiers with bi-temporal edge validity and three search methods, ElephantBroker provides comparable temporal capabilities and extends the architecture with five retrieval sources (adding structural property queries and artifact search), eleven-dimension scoring with budget competition, evidence verification, guard pipelines, goal-aware extraction, and consolidation-based learning. Zep's community subgraph, which clusters entities into summarized neighborhoods via label propagation, has no direct counterpart in ElephantBroker, though our consolidation engine's clustering and canonicalization stages (Section~\ref{subsec:consolidation}, Stages~1--2) serve a related purpose of reducing redundancy across the knowledge base. Compared with Mem0~\cite{ref-mem0-2025}, which achieves strong retrieval performance and low latency, ElephantBroker adds goal-aware scoring, evidence tracking, guard pipelines, and budget-constrained working set assembly at the cost of greater operational complexity. None of the compared systems provide enforceable guard pipelines with physical tool-call interception or AI-powered safety scanning for knowledge base contamination prevention. Similarly, none implements a numeric authority hierarchy with configurable permission rules for multi-organization access control. These capabilities are essential for enterprise deployments where regulatory compliance demands demonstrable enforcement, not advisory compliance.

\subsection{Limitations}

Several limitations should be acknowledged. The current evaluation is architectural rather than empirical: we validate correctness through systematic testing but do not yet report end-to-end task completion metrics or hallucination reduction rates on standardized benchmarks. The initial scoring weights are set through expert judgment and profile presets, with the consolidation engine's scoring tuner providing EMA-based weight adjustment from usage feedback, capped at $\pm 5\%$ per cycle to prevent oscillation. The system's reliance on external services (Neo4j, Qdrant, Redis, embedding API, reranker API) introduces operational complexity that may not be suitable for all deployment scenarios; the embedding cache and graceful degradation hierarchy mitigate but do not eliminate this concern.

Future work includes empirical evaluation on established memory benchmarks such as LongMemEval~\cite{ref-longmemeval} (which tests temporal reasoning, multi-session recall, and knowledge updates across conversations averaging 115{,}000 tokens) and the Deep Memory Retrieval task~\cite{ref-packer2024}, extension of the evidence model to support probabilistic reasoning, investigation of community-level summarization as explored by Zep~\cite{ref-zep2025} and GraphRAG~\cite{ref-edge2024}, and investigation of the consolidation engine's impact on long-term memory quality.

%%%%%%%%%%%%%%%%%%%%%%%%%%%%%%%%%%%%%%%%%%
\section{Conclusions and Future Work}
\label{sec:conclusions}

This paper argues, based on the growing demands for trustworthy AI agents and the limitations of existing memory systems, the need for a cognitive runtime that treats the trustworthiness of stored knowledge as a first-class concern in context assembly.

We presented ElephantBroker, a knowledge-grounded cognitive runtime that unifies a Neo4j knowledge graph with a Qdrant vector store through the Cognee SDK to implement a complete cognitive loop: store knowledge with typed provenance, retrieve it through hybrid five-source search with profile-driven isolation, score candidates in an eleven-dimension budget competition with verification-weighted confidence, compose context through four-block assembly across two prompt surfaces, protect agent actions through a six-layer cheap-first guard pipeline with physical enforcement via the AI Firewall, and learn from usage patterns through nine-stage consolidation. The system's eight principal contributions (hybrid retrieval, competitive scoring, evidence-tracked verification, context lifecycle management, guard-based safety enforcement, consolidation-based learning, AI Firewall enforcement, and authority-based identity) address key gaps in existing agent memory systems and together close the cognitive loop from knowledge acquisition to knowledge refinement. The AI Firewall provides dual-enforcement guard compliance through tool-call interception at three defense layers and six-tier cost-ordered safety scanning, and the authority-based identity model enables enterprise-grade multi-organization access control through a numeric permission hierarchy.

The evidence verification model, with its graded confidence multipliers linking verification state to scoring priority, creates a structural bias toward trustworthy information. The procedure--guard--evidence triangle provides enterprise-grade compliance infrastructure where procedures define workflows, guards enforce safety constraints, and evidence ensures accountability. The consolidation engine, modeled on human sleep-based memory processing, continuously strengthens useful knowledge while decaying noise, with seven of nine stages requiring zero LLM calls. These design principles, that memory systems should not merely store and retrieve but should also track trustworthiness, enforce safety, and learn from usage, represent a shift from retrieval-centric to verification-aware cognitive architectures.

ElephantBroker is released as open-source software under the AGPL-3.0 license, with a comprehensive test suite validating schemas, adapters, runtime modules, ingest pipelines, retrieval, scoring, reranking, session goal management, and procedure lifecycle tracking. In the future, we intend to conduct empirical evaluation on standardized multi-turn benchmarks to measure the impact of verification-aware scoring on hallucination reduction and task completion rates, and to investigate the consolidation engine's long-term effect on knowledge quality across extended multi-session interactions.

%%%%%%%%%%%%%%%%%%%%%%%%%%%%%%%%%%%%%%%%%%
\vspace{1em}

\noindent\textbf{Author Contributions:} Conceptualization, C.L. and A.L.; Methodology, C.L.; Software, C.L. and A.L.; Validation, C.L. and A.L.; Formal Analysis, C.L.; Investigation, C.L.; Resources, C.L.; Data Curation, C.L.; Writing---Original Draft Preparation, C.L.; Writing---Review and Editing, C.L. and A.L.; Visualization, C.L.; Supervision, C.L.; Project Administration, C.L. and A.L. All authors have read and agreed to the published version of the manuscript.

\noindent\textbf{Funding:} This research received no external funding.

\noindent\textbf{Data Availability:} The source code of the ElephantBroker system is openly available at \url{https://github.com/elephant-broker/elephant-broker} under the AGPL-3.0 license. No datasets were generated or analyzed during this study; the evaluation is based on architectural validation through the project's test suite.

\noindent\textbf{Acknowledgments:} During manuscript preparation, the authors used a generative AI tool for limited language refinement and formatting assistance. All technical content, implementation decisions, validation, and conclusions were reviewed and verified by the authors, who take full responsibility for the manuscript.

\noindent\textbf{Conflicts of Interest:} The authors are the founders of Elephant Broker. The authors declare that this affiliation did not inappropriately influence the representation or interpretation of the reported research results.

\vspace{1em}

\noindent\textbf{Abbreviations:}
The following abbreviations are used in this manuscript:

\medskip
\noindent
\begin{tabular}{@{}ll}
LLM & Large Language Model\\
RAG & Retrieval-Augmented Generation\\
API & Application Programming Interface\\
SDK & Software Development Kit\\
GDPR & General Data Protection Regulation\\
OTEL & OpenTelemetry\\
EMA & Exponential Moving Average\\
TTL & Time to Live\\
SOP & Standard Operating Procedure
\end{tabular}

%%%%%%%%%%%%%%%%%%%%%%%%%%%%%%%%%%%%%%%%%%
\bibliographystyle{unsrt}

\begin{thebibliography}{99}

\bibitem{ref-park2023}
J.~S. Park, J.~C. O'Brien, C.~J. Cai, M.~R. Morris, P.~Liang, and M.~S. Bernstein, ``Generative Agents: Interactive Simulacra of Human Behavior,'' in \textit{Proc. 36th Annual ACM Symposium on User Interface Software and Technology (UIST)}, San Francisco, CA, USA, 2023, pp.~1--22.

\bibitem{ref-sumers2024}
T.~R. Sumers, S.~Yao, K.~Narasimhan, and T.~L. Griffiths, ``Cognitive Architectures for Language Agents,'' \textit{Trans. Mach. Learn. Res.}, vol.~2024, pp.~1--45, 2024.

\bibitem{ref-lewis2020}
P.~Lewis, E.~Perez, A.~Piktus, F.~Petroni, V.~Karpukhin, N.~Goyal, H.~K\"{u}ttler, M.~Lewis, W.~Yih, T.~Rockt\"{a}schel, \textit{et~al.}, ``Retrieval-Augmented Generation for Knowledge-Intensive NLP Tasks,'' in \textit{Advances in Neural Information Processing Systems}, vol.~33, 2020, pp.~9459--9474.

\bibitem{ref-ji2023}
Z.~Ji, N.~Lee, R.~Frieske, T.~Yu, D.~Su, Y.~Xu, E.~Ishii, Y.~J. Bang, A.~Madotto, and P.~Fung, ``Survey of Hallucination in Natural Language Generation,'' \textit{ACM Comput. Surv.}, vol.~55, pp.~1--38, 2023.

\bibitem{ref-hong2024}
S.~Hong, M.~Zhuge, J.~Chen, X.~Zheng, Y.~Cheng, J.~Wang, C.~Zhang, Z.~Wang, S.~K.~S. Yau, Z.~Lin, \textit{et~al.}, ``MetaGPT: Meta Programming for A Multi-Agent Collaborative Framework,'' in \textit{Proc. 12th International Conference on Learning Representations (ICLR)}, Vienna, Austria, 2024.

\bibitem{ref-packer2024}
C.~Packer, S.~Wooders, K.~Lin, V.~Fang, S.~G. Patil, I.~Stoica, and J.~E. Gonzalez, ``MemGPT: Towards LLMs as Operating Systems,'' in \textit{Proc. 12th International Conference on Learning Representations (ICLR)}, Vienna, Austria, 2024.

\bibitem{ref-gao2024}
Y.~Gao, Y.~Xiong, X.~Gao, K.~Jia, J.~Pan, Y.~Bi, Y.~Dai, J.~Sun, and H.~Wang, ``Retrieval-Augmented Generation for Large Language Models: A Survey,'' \textit{arXiv preprint arXiv:2312.10997}, 2024.

\bibitem{ref-pan2024}
S.~Pan, L.~Luo, Y.~Wang, C.~Chen, J.~Wang, and X.~Wu, ``Unifying Large Language Models and Knowledge Graphs: A Roadmap,'' \textit{IEEE Trans. Knowl. Data Eng.}, vol.~36, pp.~3580--3599, 2024.

\bibitem{ref-edge2024}
D.~Edge, H.~Trinh, N.~Cheng, J.~Bradley, A.~Chao, A.~Mody, S.~Truitt, and J.~Larson, ``From Local to Global: A Graph RAG Approach to Query-Focused Summarization,'' \textit{arXiv preprint arXiv:2404.16130}, 2024.

\bibitem{ref-cognee}
``Cognee: Build Knowledge Graphs and Retrieval Pipelines for LLMs.'' Available: \url{https://github.com/topoteretes/cognee} (accessed Mar.~15, 2026).

\bibitem{ref-manakul2023}
P.~Manakul, A.~Liusie, and M.~J.~F. Gales, ``SelfCheckGPT: Zero-Resource Black-Box Hallucination Detection for Generative Large Language Models,'' in \textit{Proc. 2023 Conference on Empirical Methods in Natural Language Processing (EMNLP)}, Singapore, 2023, pp.~9004--9017.

\bibitem{ref-ma2024}
X.~Ma, Y.~Gong, P.~He, H.~Zhao, and N.~Duan, ``Query Rewriting in Retrieval-Augmented Large Language Models,'' in \textit{Proc. 2024 Conference on Empirical Methods in Natural Language Processing (EMNLP)}, Miami, FL, USA, 2024, pp.~5303--5315.

\bibitem{ref-trivedi2023}
H.~Trivedi, N.~Balasubramanian, T.~Khot, and A.~Sabharwal, ``Interleaving Retrieval with Chain-of-Thought Reasoning for Knowledge-Intensive Multi-Step Questions,'' in \textit{Proc. 61st Annual Meeting of the Association for Computational Linguistics (ACL)}, Toronto, ON, Canada, 2023, pp.~10014--10037.

\bibitem{ref-asai2024}
A.~Asai, Z.~Wu, Y.~Wang, A.~Sil, and H.~Hajishirzi, ``Self-RAG: Learning to Retrieve, Generate, and Critique through Self-Reflection,'' in \textit{Proc. 12th International Conference on Learning Representations (ICLR)}, Vienna, Austria, 2024.

\bibitem{ref-laird2012}
J.~E. Laird, \textit{The Soar Cognitive Architecture}. Cambridge, MA, USA: MIT Press, 2012.

\bibitem{ref-anderson2004}
J.~R. Anderson, D.~Bothell, M.~D. Byrne, S.~Douglass, C.~Lebiere, and Y.~Qin, ``An Integrated Theory of the Mind,'' \textit{Psychol. Rev.}, vol.~111, pp.~1036--1060, 2004.

\bibitem{ref-huang2023}
L.~Huang, W.~Yu, W.~Ma, W.~Zhong, Z.~Feng, H.~Wang, Q.~Chen, W.~Peng, X.~Feng, B.~Qin, \textit{et~al.}, ``A Survey on Hallucination in Large Language Models: Principles, Taxonomy, Challenges, and Open Questions,'' \textit{arXiv preprint arXiv:2311.05232}, 2023.

\bibitem{ref-min2023}
S.~Min, K.~Krishna, X.~Lyu, M.~Lewis, W.~Yih, P.~W. Koh, M.~Iyyer, L.~Zettlemoyer, and H.~Hajishirzi, ``FActScore: Fine-grained Atomic Evaluation of Factual Precision in Long Form Text Generation,'' in \textit{Proc. 2023 Conference on Empirical Methods in Natural Language Processing (EMNLP)}, Singapore, 2023, pp.~12076--12100.

\bibitem{ref-dhuliawala2023}
S.~Dhuliawala, M.~Komeili, J.~Xu, R.~Raileanu, X.~Li, A.~Celikyilmaz, and J.~Weston, ``Chain-of-Verification Reduces Hallucination in Large Language Models,'' \textit{arXiv preprint arXiv:2309.11495}, 2023.

\bibitem{ref-kadavath2022}
S.~Kadavath, T.~Conerly, A.~Askell, T.~Henighan, D.~Drain, E.~Perez, N.~Schiefer, Z.~Hatfield-Dodds, N.~DasSarma, E.~Tran-Johnson, \textit{et~al.}, ``Language Models (Mostly) Know What They Know,'' \textit{arXiv preprint arXiv:2207.05221}, 2022.

\bibitem{ref-euaiact}
European Parliament and Council, ``Regulation (EU) 2024/1689 Laying Down Harmonised Rules on Artificial Intelligence (Artificial Intelligence Act),'' \textit{Off. J. Eur. Union}, L 2024/1689, pp.~1--144, 2024.

\bibitem{ref-luo2024}
L.~Luo, Y.~F. Li, G.~Haffari, and S.~Pan, ``Reasoning on Graphs: Faithful and Interpretable Large Language Model Reasoning,'' in \textit{Proc. 12th International Conference on Learning Representations (ICLR)}, Vienna, Austria, 2024.

\bibitem{ref-shinn2023}
N.~Shinn, F.~Cassano, A.~Gopinath, K.~Narasimhan, and S.~Yao, ``Reflexion: Language Agents with Verbal Reinforcement Learning,'' in \textit{Advances in Neural Information Processing Systems}, vol.~36, 2023, pp.~8634--8652.

\bibitem{ref-zep2025}
P.~Rasmussen, P.~Paliychuk, T.~Beauvais, J.~Ryan, and D.~Chalef, ``Zep: A Temporal Knowledge Graph Architecture for Agent Memory,'' \textit{arXiv preprint arXiv:2501.13956}, 2025.

\bibitem{ref-mem0-2025}
P.~Chhikara, D.~Khant, T.~Arora, D.~Singh, and T.~Bindra, ``Mem0: Building Production-Ready AI Agents with Scalable Long-Term Memory,'' \textit{arXiv preprint arXiv:2504.19413}, 2025.

\bibitem{ref-losslessclaw}
``Lossless-Claw: Lossless Context Compression for OpenClaw Agents.'' Available: \url{https://github.com/Martian-Engineering/lossless-claw} (accessed Mar.~15, 2026).

\bibitem{ref-supermemory}
``Supermemory: Cloud Memory API for AI Agents.'' Available: \url{https://supermemory.ai} (accessed Mar.~15, 2026).

\bibitem{ref-qmd}
``QMD: Local Hybrid Search Sidecar for LLM Agents.'' Available: \url{https://github.com/tobi/qmd} (accessed Mar.~15, 2026).

\bibitem{ref-nemhauser1978}
G.~L. Nemhauser, L.~A. Wolsey, and M.~L. Fisher, ``An Analysis of Approximations for Maximizing Submodular Set Functions---I,'' \textit{Math. Program.}, vol.~14, pp.~265--294, 1978.

\bibitem{ref-longmemeval}
D.~Wang, H.~Zhu, D.~Yang, Y.~Lin, K.~Qian, H.~Yao, X.~Li, J.~X. Huang, and D.~Yu, ``LongMemEval: Benchmarking Chat Assistants on Long-Term Interactive Memory,'' \textit{arXiv preprint arXiv:2410.10813}, 2024.

\bibitem{ref-owasp-llm}
OWASP Foundation, ``OWASP Top 10 for Large Language Model Applications, Version 2.0,'' 2025. Available: \url{https://owasp.org/www-project-top-10-for-large-language-model-applications/} (accessed Mar.~24, 2026).

\bibitem{ref-greshake2023}
K.~Greshake, S.~Abdelnabi, S.~Mishra, C.~Endres, T.~Holz, and M.~Fritz, ``Not What You've Signed Up For: Compromising Real-World LLM-Integrated Applications with Indirect Prompt Injection,'' in \textit{Proc. AISec Workshop at ACM CCS}, Copenhagen, Denmark, 2023, pp.~79--90.

\bibitem{ref-perez2023}
F.~Perez and I.~Ribeiro, ``Ignore This Title and HackAPrompt: Exposing Systemic Weaknesses of Language Models via a Global Scale Prompt Hacking Competition,'' in \textit{Proc. 2023 Conference on Empirical Methods in Natural Language Processing (EMNLP)}, Singapore, 2023, pp.~4945--4977.

\bibitem{ref-llmguard}
ProtectAI, ``LLM Guard: Security Toolkit for LLM Interactions.'' Available: \url{https://github.com/protectai/llm-guard} (accessed Mar.~24, 2026).

\bibitem{ref-guardrailsai}
Guardrails AI, ``Guardrails: Adding Guardrails to Large Language Models.'' Available: \url{https://github.com/guardrails-ai/guardrails} (accessed Mar.~24, 2026).

\bibitem{ref-nemo}
T.~Rebedea, R.~Dinu, M.~Sreedhar, C.~Parisien, and J.~Cohen, ``NeMo Guardrails: A Toolkit for Controllable and Safe LLM Applications with Programmable Rails,'' in \textit{Proc. 2023 Conference on Empirical Methods in Natural Language Processing: System Demonstrations (EMNLP)}, Singapore, 2023, pp.~431--445.

\bibitem{ref-rebuff}
ProtectAI, ``Rebuff: Prompt Injection Detector.'' Available: \url{https://github.com/protectai/rebuff} (accessed Mar.~24, 2026).

\bibitem{ref-litellm}
BerriAI, ``LiteLLM: Call 100+ LLM APIs in the OpenAI Format.'' Available: \url{https://github.com/BerriAI/litellm} (accessed Mar.~24, 2026).

\bibitem{ref-dong2024}
Y.~Dong, R.~Jiang, H.~Jin, Y.~Lin, H.~Yang, Y.~Zheng, Z.~Wang, L.~Sun, B.~Peng, J.~Deng, \textit{et~al.}, ``Safeguarding Large Language Models: A Survey,'' \textit{arXiv preprint arXiv:2406.02622}, 2024.

\bibitem{ref-nist-ai}
National Institute of Standards and Technology (NIST), ``Artificial Intelligence Risk Management Framework (AI RMF 1.0),'' NIST AI 100-1, Gaithersburg, MD, USA, 2023.

\end{thebibliography}

\end{document}